\documentclass[journal]{IEEEtran}

\usepackage[utf8]{inputenc}
\usepackage{graphicx}
\usepackage{color}
\usepackage{amsfonts}
\usepackage{amsmath}
\usepackage[switch]{lineno}
\usepackage[pagebackref=true,breaklinks=true,colorlinks,bookmarks=false]{hyperref}
\usepackage{booktabs}
\usepackage{caption} % caption centering
\usepackage{subcaption}
\usepackage[export]{adjustbox} % align table and figure
\usepackage{cancel}
\usepackage{cite}
\usepackage{multirow}
\usepackage{algorithm}
\usepackage{algorithmic}

\usepackage[resetlabels,labeled]{multibib}
\newcites{Ref}{references}

\newcommand{\edit}[1]{#1}
\newcommand{\add}[1]{#1}

\title{Towards Robust Neural Image Compression: Adversarial Attack and  Model Finetuning}

\author{Tong~Chen,~\IEEEmembership{Student Member,~IEEE,} and~Zhan~Ma,~\IEEEmembership{Senior Member,~IEEE}% <-this % stops a space
\thanks{This work was partially supported by the National Natural Science Foundation of China under Grant 62022038.}
\thanks{Copyright © 2023 IEEE. Personal use of this material is permitted. However, permission to use this material for any other purposes must be obtained from the IEEE by sending an email to pubs-permissions@ieee.org.}
\thanks{T. Chen and Z. Ma are with the School of Electronic Science and Engineering, Nanjing University, Nanjing, Jiangsu, 210093 China. E-mails: tong@smail.nju.edu.cn, mazhan@nju.edu.cn. (Corresponding Author: Z. Ma)}% <-this % stops a space
}
\begin{document}
%\linenumbers
\maketitle
\begin{abstract}
Deep neural network-based image compression has been extensively studied. However, the model robustness  which is crucial to practical application is largely overlooked.
We propose to examine the robustness of prevailing learned image compression models by injecting negligible adversarial perturbation into the original source image. Severe distortion in decoded reconstruction reveals the general vulnerability in existing methods regardless of their  settings (e.g., network architecture, loss function, quality scale).
A variety of defense strategies including geometric self-ensemble based pre-processing, and adversarial training, are investigated against the adversarial attack to improve the model's robustness. Later the defense efficiency is further exemplified in real-life image recompression case studies.
%\sout{We then explore possible defense strategies against the adversarial attack to improve the model robustness, including geometric self-ensemble based pre-processesing, and adversarial training. Experiments report the effectiveness of various defense strategies. Additional image recompression case study further confirms the substantial improvement of the robustness of compression models in real-life applications.} 
Overall, our methodology is simple, effective, and generalizable, making it attractive for developing robust learned image compression solutions.
%\edit{Adversarial attacks have shown the ability to misguide the neural networks with incorrect output}. In this paper, we present both targeted and untargeted adversarial attack strategy~\gpynote{(give more specific illustration to assist understanding attack)} for learned neural image compression methods, degrading both objectively and subjectively the quality of the reconstructed images. To our knowledge, this is the first work that systematically evaluates the robustness of neural image compression. Experiments are performed on many existing methods, quality metrics as well as bitrate targets, showing that most learned image compression methods are vulnerable to adversarial attacks. Finally, an adversarial data augmentation strategy for refining the image compression network is proposed, which is able to improve the robustness of learned image compression codecs.
All materials are made publicly accessible at \url{https://njuvision.github.io/RobustNIC} for reproducible research.
\end{abstract}


\section{Introduction}
\label{sec:intro}
In the past few years, we have witnessed an exponential growth of deep neural network (DNN) based image coding approaches. These neural network-based image coding (NIC) methods~\cite{balle2016end,balle2018variational,li2018learning,minnen2018joint,mentzer2020high,Cheng2020Learned,chen2020,hu2021learning,9810760,nnc_overview} have emerged  with noticeable compression gains over conventional solutions like JPEG~\cite{wallace1991jpeg}, JPEG 2000~\cite{JPEG2K}, and even the latest Versatile Video Coding (VVC) based Intra Profile (VVC Intra)~\cite{vvc_overview}, promising an encouraging prospects of NIC-based applications and services, particularly for those rate-constrained or bandwidth-limited networked scenarios. 

Unprecedented coding gains of aforementioned NIC methods mostly came from the introduction of variational autoencoder (VAE) framework~\cite{balle2018variational}, adaptive context modeling in entropy coding~\cite{minnen2018joint}, attention mechanism~\cite{chen2020,Cheng2020Learned}, etc. It is also worth pointing out that learning-based NIC offered great flexibility for promptly supporting different image sources (e.g., RGB, YUV, Bayer raw image, etc~\cite{Zhihao_RAW_Coding}) and loss functions (quality metrics)~\cite{MS-SSIM,wang2021subjective} in task-oriented applications. On the contrary, it requires substantial efforts to enable additional functionalities in traditional image codecs because we have to fully understand the statistical properties of newly-introduced image sources and optimization metrics to develop effective coding toolkits. For example, it takes several years to facilitate the efficient compression of screen content sources when standardizing the extensions of High-Efficiency Video Coding (HEVC)~\cite{screen2016}.

% \begin{figure}[t]
% \centering
% \includegraphics[scale=0.3]{./figs/fidelity.pdf}
% \caption{Reconstruction fidelity illustration of close-up patches extracted from conventional JPEG and learning-based NIC compressed images.
% All patches are from the same location marked using red box in original image ({\it Upper Part}). (a) uncompressed patch from original input, (b) reconstructed patch of JPEG compressed image (\url{http://libjpeg.sourceforge.net/}), and (c-e) are the reconstructed patches from the same image compressed using MSE (Mean Squared Error), MS-SSIM (Multiscale Structural Similarity)~\cite{MS-SSIM}, LPIPS (Learned Perceptual Image Patch Similarity)~\cite{zhang2018unreasonable} optimized NIC methods. All of them are compressed with close PSNR (Peak Signal-to-Noise Ratio) at {about 24} dB  for fair comparison.}
% \label{fidelity}
% \end{figure}

Recently, after serial exploration studies, the ISO/IEC JPEG (Joint Photographic Experts Group) committee has made concrete progress on the performance and complexity evaluation of various NIC solutions and is now standardizing the next-generation image codec using deep learning techniques. 
%\delete{Note that deep learning is now already widely applied for the enabling of artificial intelligence (AI) in many aspects, particularly for visual computing-based services (such as the facial recognition). Thus this new learning-based image coding approach under JPEG is referred to as the ``JPEG AI''.} 
%There also merge some works trying to integrate the segmentation network with image compression network in a fully end-to-end learnable framework for object-based image coding~\cite{xia2020object}, and joint compression and computer vision task~\cite{liu2019codedretrieval}. \edit{The interaction between image compression and vision tasks has also attracted some attention.} Dziugaite et al.~\cite{dziugaite2016study} found that JPEG compression method can improve a large number of network model recognition accuracy decline caused by Fast Gradient Sign Method (FGSM)~\cite{43405} attack disturbance. 
% Targeting for the international standard for the interoperability enabling, 
Besides the superior compression efficiency, model robustness/generalization is yet another critical factor for the potential success of such learned approaches, which is often overlooked and not thoroughly examined in the past. %Therefore, this work specifically pays attention on the model robustness and generalization of recently emerged NIC approaches. 
% extend to full resolution...
%  Adversarial 
% Image resolution, brightness and contrast is 
% vulnerability.
%\gpynote{G: The above paragraphs want to highlight the importance of learning-based image compression methods. But these words doesn't powerfully introduce the key problem or obvious performance improvement that learning-based methods could solve while conditional methods couldn't. The introduction of JPEG AI may not need a whole paragraph.}

\begin{figure}[t]
    \centering
\includegraphics[width=0.49\textwidth]{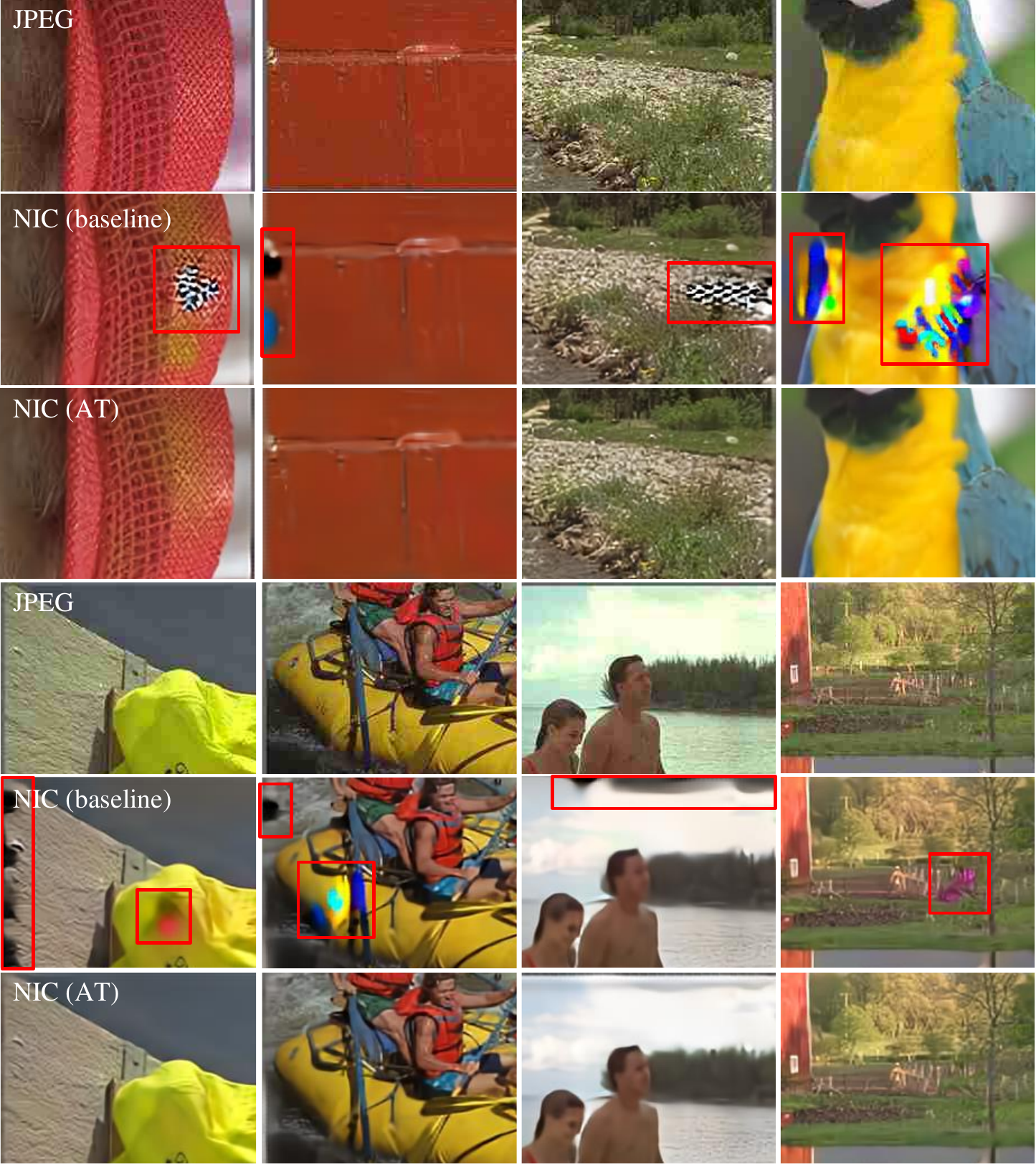}
    \caption{{\bf Image Recompression.} Row-wisely visual comparison of image reconstructions after being repetitively compressed 50 times using the JPEG (\url{http://libjpeg.sourceforge.net/}), a pretrained NIC baseline model (Ball\'e2018~\cite{balle2018variational} is used as an example), and an adversarially retrained Ball\'e2018 model, referred to as NIC (AT). More details can be found in Sec.~\ref{sec:attack_defense}.}
    \label{fig:recompression}
\end{figure}

{\bf Observation and Motivation.} 
%{Though rules-based conventional lossy image coding methods, such as the JPEG or VVC Intra would inevitably introduce compression artifacts, yielding visually unpleasant blurring, blocking, and ringing impairments in decoded reconstructions. These artifacts mainly deteriorate pixel intensity, whilst the local structure of decoded spatial texture, e.g., edge orientations, are mostly preserved as the input source.  However, existing learned image compression approaches are usually trained and tested on limited datasets. Their performance over a broad range of real-world application scenarios has not been fully validated.}
We first present a real-life application example in Fig.~\ref{fig:recompression}. As seen, unexpected pixel impairments are observed if the image is compressed multiple times using learned approaches, {while similar phenomena are not detected in images  re-compressed using rules-based conventional  methods like JPEG.}  Such iterative or successive compression is a common practice for image-based applications over the Internet where images may be re-edited, re-compressed, and re-distributed often. Similar observations are also reported by Kim {\it et. al} in~\cite{kim2020}.

Nevertheless, such instability may not be solely in image recompression scenarios~\cite{kim2020}.  A systematic stability and generalization study of popular VAE-based compression models on full-resolution images still remains untouched and is urgently needed for the pursuit of robust NIC methods because
{the signal fidelity of image reconstruction plays a vital role in vast applications} such as medical imaging, autonomous self-driving, face recognition, security surveillance, etc. For instance, having incorrect (unexpected) spatial textures in decoded images may lead to very different outcomes of specific tasks (e.g., image classification)~\cite{choi2020task} (see Fig.~\ref{mnist_comparison} and Fig.~\ref{license} in the supplemental material), or may also present annoying/disturbing artifacts~\cite{kim2020} with severely-degraded user experience (see Fig.~\ref{fig:example}).

\begin{figure}[t]
    \centering
    \includegraphics[width=0.49\textwidth]{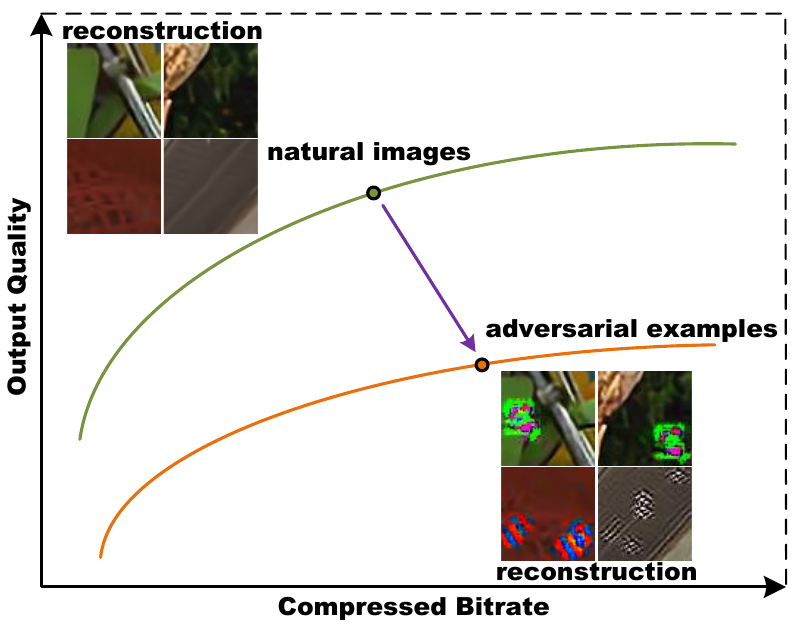}
\caption{\add{\textbf{Impact of adversarial attack in image compression.} Adversarial examples are generated to test whether the overall coding performance of NIC models would be affected.}}
    \label{fig:objective}
\end{figure}

Technically, an image compression network sometimes might be roughly categorized as one type of image-to-image task model. However, we would like to note that there are fundamental differences between image compression and popular image-to-image tasks. A typical image-to-image processing task mainly optimizes the distance (distortion) between the output target and ground truth, while image compression deals with the rate and distortion jointly. 

Generally, an image compression network separates the encoder and decoder, and applies the quantization   upon the activation elements at the bottleneck layer for rate-distortion optimization. %according to the entropy constraint
%Extra  and quantization are applied that connects these two parts. %The characteristics of the bottleneck layer are well worth investigating.
Various quantization levels would affect the sparsity of the latent representation and finally result in different quality scales (or target bitrates). As illustrated in Fig.~\ref{fig:objective}, we try to generate adversarial examples by adding negligible perturbation into the original source images, but the compression of adversarial examples would lead to severe distortion in decoded reconstruction. 
As seen, the robustness of an image compression algorithm has affected the reconstruction quality and compression bitrate jointly. 
%In this paper, we mainly focus on the attack and defense for quality degradation, but their impact on the compressed bitrate is also observed and analyzed.}

 %Besides, a variety of rate-distortion tradeoffs are commonly adapted in image compression to meet practical constraints (e.g., network bandwidth), with which  different bitrates (or quality scales) .
%In order to get an optimal rate-distortion trade-off, the image compression networks care not only about the reconstruction quality but also about the bitrate of the latent features. Therefore, extra entropy constraint and quantization are applied to the activation of the bottleneck layer to ensure the sparsity of the representation. 
%Besides, a set of hyperparameters $\lambda$s are applied to adapt trade-offs between the rate and the distortion, resulting in separate models for different bitrates (or quality scales).
All of these thus make the robustness analysis of existing image compression networks an issue worthy of a separate and in-depth study.

{\bf Approach.} 
%The aforementioned attempt in~\cite{kim2020} focused on the instability of successive deep image compression. 
Unfortunately, it is almost impossible to manually search for all potential cases that may cause model instability and deteriorate performance.
% \edit{As inspired by emerged popular adversarial attack studies on high-level vision tasks like classification~\cite{AdvExp_Goodfellow,43405,nguyen2015deep}, semantic segmentation~\cite{Xiao_2018_ECCV} and object detection~\cite{ijcai2019-134}, and low-level tasks like super resolution~\cite{yin2018sr} and optical flow derivation~\cite{ranjan2019attacking} to respectively examine their model robustness,}
As inspired by recent adversarial attack studies on high-level vision tasks~\cite{AdvExp_Goodfellow,43405,nguyen2015deep,Xiao_2018_ECCV,ijcai2019-134} and low-level image-to-image tasks~\cite{yin2018sr,ranjan2019attacking,choi_deep_2021} to respectively examine their model robustness,
we suggest implementing the adversarial attack to analyze and characterize its impacts on the robustness and coding efficiency of the underlying NIC models, quantitatively and qualitatively. 

% \add{Note that there have been some previous works~\cite{yin2018sr,ranjan2019attacking,choi_deep_2021} on attacking image-to-image networks. \edit{Successful attacks on these networks indicate that image compression networks are also very likely to be vulnerable to adversarial attacks.
% However, there are some major differences between image compression and other image processing networks.}

Later we extensively explore possible defense strategies and their effects on the robustness improvement of NIC methods to adversarial attacks. An implementation of our defense strategy in practical application is also exemplified. As shown in Fig.~\ref{fig:recompression} and Fig.~\ref{fig:recompression_rd}, the proposed defense strategies can effectively eliminate the distortions induced by image recompression. 

{\bf Contributions.}
\begin{enumerate}
    \item To the best of our knowledge, this is the first work that performs the adversarial attack to systematically study the model robustness of learned image compression on full resolution images (Sec.~\ref{sec:method}); Different from most existing works that mainly qualitatively evaluate the attack and defense methods, we propose a metric -- $\Delta$PSNR for quantitative analysis which is shown to be effective in experiments. 
    \item Our extensive experiments report the general vulnerability of existing learning-based image compression methods regardless of the underlying model settings and the perturbation generation strategy 
    (Sec.~\ref{sec:exp}); A novel Fast Threshold-constrained Distortion Attack (FTDA) approach is proposed to generate adversarial examples with balanced performance and complexity while existing methods like CW~\cite{carlini_towards_2017} or I-FGSM~\cite{kurakin_adversarial_2017} mostly satisfy one aspect, i.e., performance or complexity.
    %much better attack performance than the Fast gradient-based techniques like I-FGSM~\cite{}  
    \item We further improve the model's robustness against attacks through the use of various defense strategies including geometric self-ensemble, adversarial training as well as image pre-processing techniques like spatial resizing and bit depth reduction. Their impact on compression efficiency is thoroughly examined and discussed (Sec.~\ref{sec:attack_defense}). 
    As a result, adversarial training and geometric self-ensemble are suggested because of their convincing efficiency.
    while spatial resizing or bit depth reduction induced pre-processing distortion inherently deteriorates the performance of the underlying compression model which is not acceptable for applications. 
      
 %(Sec.~\ref{sec:attack_defense}). 
    %\item \add{Sec.~\ref{sec:targeted}}
\end{enumerate}
 Overall, our methodology which includes the adversarial example generation and the defense strategies for improving the model robustness is simple, effective, and generalizable to  popular learned image compression approaches, making it attractive and necessarily desired for practical application.

\section{Related Works}
\label{sec:related_work}
This section briefs relevant techniques on respective lossy image compression and adversarial attack.

\begin{figure*}
\begin{subfigure}[b]{0.5\textwidth}
\centering
\includegraphics[scale=0.75]{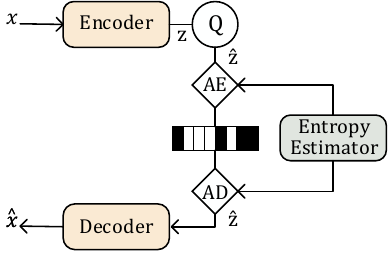}
\caption{}
\label{fig:vae}
\end{subfigure}%
\begin{subfigure}[b]{0.5\textwidth}
\centering
\begin{tabular}{c|c}
     \textbf{Encoder} & \textbf{Decoder} \\ \midrule
     Conv/Resblock s2$\downarrow$ & Deconv/Resblock s2$\uparrow$ \\
     GDN/ReLU     & IGDN/ReLU      \\
     Conv/Resblock s2$\downarrow$ & Deconv/Resblock s2$\uparrow$ \\
     GDN/ReLU     & IGDN/ReLU      \\
     Conv/Resblock s2$\downarrow$ & Deconv/Resblock s2$\uparrow$ \\
     GDN/ReLU     & IGDN/ReLU      \\
     Conv/Resblock s2$\downarrow$ & Deconv/Resblock s2$\uparrow$ \\ 
\end{tabular}
\caption{}
\label{table:vae}
\end{subfigure}
\caption{{\bf Neural Image Compression (NIC).} (a) \add{Basic} architecture of VAE-based NIC Solutions. ${\mathbf{Q}}$ denotes the quantization. The entropy estimator provides the probability estimation for both AE (Arithmetic Encoding) and AD (Arithmetic Decoding); $\bf x$ is the input source image, and correspondingly $\bf \hat{x}$ is the output reconstruction (after decoding). (b) Typical paired Encoder and Decoder components used in NIC framework, such as normal Convolutional layer (Conv) or ResNet block (Resblock)~\cite{he2016deep}, resolution re-sampling (upscaling s2$\uparrow$ and downscaling s2$\downarrow$ at a factor of 2), nonlinear activations using GDN (Generalized Divisive Normalization)~\cite{balle2015density} or ReLU (rectified linear unit)~\cite{nair2010rectified}. Deconv and IGDN are inverted processes of respective Conv and GDN.} 
\end{figure*}

\subsection{Lossy Image Compression}\label{sec:related_lossy}
Lossy image compression approaches like JPEG~\cite{wallace1991jpeg}, JPEG 2000~\cite{JPEG2K}, VVC Intra~\cite{vvc_overview}, etc,  search for appropriate compact representation through rate-distortion optimization (RDO).  Typically, they adopt the ``transform coding'' or ``hybrid transform/prediction coding'' to transform input pixels, or pixel residuals (after intra prediction) into frequency domain coefficients for quantization and entropy coding. These transforms are generally comprised of a set of basis functions that are weighted to represent arbitrary pixel-domain blocks/patches using a few sparse, nonzero coefficients.

{\bf Classical Methods.} Classical transforms include Discrete Cosine Transform (DCT)~\cite{DCT}, Wavelet Transform~\cite{J2K_Wavelet}, and so on. Taking DCT as an example, transformed coefficients are usually clustered at a few low-frequency positions which can be leveraged for energy compaction. The DCT is invertible, i.e., the original image $\bf x$ can be reconstructed losslessly by applying the inverse DCT (IDCT), e.g., ${\bf x}=f^{-1}_{\text{dct}}(f_{\text{dct}}({\bf x}))$. Here $f_{\text{dct}}$ and $f^{-1}_{\text{dct}}$ are DCT and IDCT respectively. 

{\bf Over-complete Dictionary.} Image data typically exhibit non-stationary behavior, making it difficult to be de-correlated sufficiently by existing DCT or wavelets.  Dictionary learning is then introduced to use a number of content-adaptive, over-complete dictionary atoms to represent image blocks using more sparse and easy-to-compress coefficients~\cite{zepeda2011image,Yao_SelfAdaptive}.

%Classical DCT and Wavelets are mathematically explainable

%Often times, lossy compression is achieved by quantizing transform coefficients. The optimal quantizer is determined by optimizing the rate-distortion cost. In principle,  larger quantization introduces more visible artifacts (e.g., blurry appearance) and higher distortion.

%of original image or prediction error image followed by an adaptive entropy coding. During this procedure, all distortion arise from the quantization function of the transform coefficients. High-frequency texture and details are suppressed. As a result, the reconstructed images are often smoothed or blurred. 

%\edit{Oneexhibitnt reason for using these transforms is that they are mathematically explainable}, which means that they can produce reconstruction with stable quality and usually predictable distortion.
% \begin{figure}[t]
% \centering
% \includegraphics[scale=0.7]{./figs/vae_main.pdf}
% \caption{Typical Encoder/Decoder pair of VAE-based image compression framework.}
% \label{main_coder}
% \end{figure}

{\bf Neural Transform.} Built upon the excessive representation capacity of high-dimensional data, deeply stacked convolutional neural networks (CNNs) are therefore utilized to construct neural transforms to represent image blocks in a more compact means (see Fig.~\ref{table:vae}). Figure~\ref{fig:vae} plots a popular {VAE} architecture that applies neural transform and  demonstrates superior coding efficiency~\cite{balle2016end,balle2018variational,minnen2018joint,lee2018context,chen2020,Cheng2020Learned}. 

% \begin{table}[t]
%     \centering
%     \caption{ Frequently Used  Abbreviations}
%     \label{tab:notations}
%     \begin{tabular}{c|c}
%     \hline
%     Abbreviation & Description\\
%     \hline
%     NIC     & Neural Image Compression \\
%     JPEG & Joint Photographic Experts Group\\
%     VVC & Versatile Video Coding\\
%     \hline
%     MSE & Mean Squared Error\\
%     PSNR & Peak Signal-to-Noise Ratio\\
%     MS-SSIM~\cite{MS-SSIM} & Multscale Structural Similarity\\
%     LPIPS~\cite{zhang2018unreasonable} & Learned Perceptual Image Patch Similarity\\
%     bpp & bit rate using bits per pixel\\
%     GAN & Generative Adversarial Network\\
%     VAE & Variational AutoEncoder\\
%     AT & Adversarial Training\\
%     %Conv & Convolutional layer\\
%    % Resblock & ResNet~\cite{he2016deep} Blocks \\
%     \hline
%     \end{tabular}
%     %\caption{Caption}
%     %\label{tab:my_label}
% \end{table}

%use a popular VAE~\cite{kingma2013auto}-based framework with extra entropy estimation modules as demonstrated in Fig.~\ref{fig:vae}. It applies a variational autoencoder pair similar to the architecture listed in Table~\ref{table:vae}. 

The forward transform in the encoder analyzes the original image source $\bf x$ and represents it using compact, vectorized latent features $\bf z$ for quantization (e.g., $\hat{\bf z} = {\mathbf{Q}}({\bf z})$) and entropy coding with learned adaptive contexts. Oppositely, the backward transform in the decoder mirrors the forward steps in the encoding process to generate the reconstruction $\hat{\bf x}$. As usual,  the RDO module controls the compression trade-off by a hyperparameter $\lambda$, e.g.,
\begin{equation}
R + \lambda D = \underbrace{\mathbb{E}[-\log_2p({\bf \hat{z}})]}_{rate}+\lambda\underbrace{\mathbb{E}||{\bf x}-\hat{\bf x}||_2^2}_{distortion}.
\label{eq:vae_rd}
\end{equation} with $p({\bf\hat{z}})$ standing for the estimated probability of feature elements at the bottleneck layer~\cite{chen2020}.

\begin{table}[t]
\centering
\caption{Quality Scales}
\begin{tabular}{@{}lllllll@{}}
\toprule
\textbf{Quality} & \multicolumn{1}{c}{\textbf{1}} & \multicolumn{1}{c}{\textbf{2}} & \multicolumn{1}{c}{\textbf{3}} & \multicolumn{1}{c}{\textbf{4}} & \multicolumn{1}{c}{\textbf{5}} & \multicolumn{1}{c}{\textbf{6}} \\ \midrule
MSE                                  & 0.0018                         & 0.0035                         & 0.0067                         & 0.0130                         & 0.0250                         & 0.0483 \\
MS-SSIM & 2.40 & 4.58 & 8.73 & 16.64 & 31.73 & 60.50\\ \bottomrule
\label{tab:quality}
\end{tabular}
\end{table}

Unlike DCT or wavelet transform, the neural transforms $f_{{E}}()$ in the encoder and $f_{{D}}()$ in the decoder are usually non-linear functions. \add{Although formally the encoder and decoder are of symmetrical structure, their coefficients are independently random-initialized and then optimized by specific stochastic gradient descent algorithms, thus their invertibility is not mathematically guaranteed even without quantization error, i.e.,}
\begin{equation}
    {\bf x} \neq f_{{D}}(f_{{E}}({\bf x})).
    \label{eq:irreversible_neural_tramsform}
\end{equation}

Even having invertible neural transform as studied in~\cite{helminger2021lossy,xie2021enhanced}, the quantization error in lossy compression may also be amplified during the layer-by-layer decoding process. For either case, $f_D()$ and $f_E()$ are made up of a stack of convolutional layers that consist of linear convolution, non-linear activation, and optional spatial re-sampling, with which it would potentially amplify specific perturbation through layer-by-layer computation and finally lead to unexpected impairments. 

As explained in previous works~\cite{AdvExp_Goodfellow, Moosavi-Dezfooli_2016_CVPR}, nonlinear neural networks are still too linear to resist linear adversarial perturbation. By iteratively updating the adversarial perturbation with smaller steps, linear adversarial perturbation in high-dimensional spaces is sufficient to cause the vulnerability issue of underlying deep neural networks.
A successful attack to image-to-image networks also implies that deep neural transform based NIC methods are potentially vulnerable to adversarial perturbation.

%This kind of design has the advantage that we can simply use different quality metrics as objective function to get correspondingly optimized transformation parameters in an end-to-end learning manner. However, it also means that the autoencoder is a black box. It's unclear to the users that what kind of distortion would it introduce. 

%\add{Instead of using typical autoencoder style convolution networks, recently Invertible Neural Networks (INN) have been used for image compression tasks~\cite{helminger2021lossy}. Such arch itecture is composed of invertible transformations, which can eliminate the shortcuts of typical convolution layers mentioned before but the overall performance was not yet comparable with state-of-the-art VAE-based method. More recently, Xie~\cite{xie2021enhanced} further leverage a feature enhancement module to improve the nonlinear representation capacity of INN-based network.}
\begin{table*}[t]
    \centering
\caption{Tested NIC Approaches and Their Key Components.}
    \label{tab:methods}
    \begin{tabular}{l|l|l|l}
    \hline
        \multirow{2}{*}{\textbf{Methods}} & \multicolumn{3}{c}{\textbf{Key Components}} \\
        \cline{2-4}
        &  {\bf Nonlinear Transform} & \textbf{Entropy Model} & \textbf{Loss} \\
        \hline
        Ball\'e 2016~\cite{balle2016end} & Conv+GDN & factorized & MSE / MS-SSIM \\
        \hline
        Ball\'e 2018~\cite{balle2018variational} & Conv+GDN  & hyperprior & MSE / MS-SSIM \\
        \hline
        Minnen~\cite{minnen2018joint} & Conv+GDN  & joint hyperprior \& autoregressive & MSE \\
        \hline
        Cheng 2020~\cite{Cheng2020Learned} &  Resblock+Spatial-Channel Attention & joint hyperprior \&  autoregressive & MSE \\ 
        \hline
        NLAIC~\cite{chen2020} & Resblock+Nonlocal Attention & joint hyperprior \& autoregressive & MS-SSIM \\
        \hline
        HiFiC~\cite{mentzer2020high} & Conv+Channel Normalization & hyperprior & MSE + LPIPS~\cite{zhang2018unreasonable} + GAN \\
        \hline
        Weixin 2021~\cite{weixin2021fix} & Fixed-point NLAIC & joint hyperprior \& autoregressive & MSE \\
        \hline
        InvCompress~\cite{xie2021enhanced} & INN+Attention+Feature Enhancement & joint hyperprior \& autoregressive & MSE \\
        \hline
    \end{tabular}
\end{table*}

\subsection{Adversarial Attack and Defense}

{\bf Attack.} Adversarial attacks have been widely utilized to evaluate the robustness of DNN models in various approaches, particularly for those high-level vision tasks as aforementioned. 
As known, these DNN models basically wish to construct a non-linear mapping function $g$ to effectively characterize the relationships between input data and output results. 

An adversarial attack can be either \textit{targeted} or \textit{untargeted}. Given a vectorized input $\bf x$ and a mapping function $y = g({\bf x})$, 
% \edit{adversarial examples ${\bf x}^{\ast}$ that are used to perform  the \textit{untargeted} attack to produce undesignated pixel distortion in decoded images (or adversarial reconstructions)}
for an \textit{untargeted} attack, adversarial examples ${\bf x}^{\ast}$ are used to produce undesignated but incorrect output. They can be derived using: 
\begin{align}
\mathop{\arg\!\min}_{{\bf x}^{\ast}} {d}\{{\bf x},{\bf x}^{\ast}\}  \ \text{s.t.} \ g({\bf x}^{\ast}) \neq g({\bf x}). \label{eq:untargeted_attack}
\end{align} Here ${d}\{\cdot\}$ measures the signal distance between original $\bf x$ and adversarial example ${\bf x}^{\ast}$, for which conventional MSE or other metrics can be used. 
%Many popular methods like stochastic gradient decent (SGD), Adam~\cite{adam} and fast gradient-based techniques can be used for solving \eqref{eq:untargeted_attack}.% to generate respective adversarial examples to measure the robustness of neural models.

%for different purpose. 
The adversarial attack can be also {\it targeted}. Taking image classification as an example, a targeted attack, as the name implies, attempts to use example ${\bf x}^*$ that has limited perturbation compared with original input $\bf x$, to “fool” the classifier $f$ to misclassify ${\bf x}^*$ to some targeted label $y^t = f({\bf x}^*)$ that is different from the original outcome $y$. Such targeted adversarial examples  can be  generated through: 
\begin{align}
\mathop{\arg\!\min}_{{\bf x}^{\ast}} {d}\{{\bf x},{\bf x}^{\ast}\} \ \text{s.t.} \ f({\bf x}^{\ast}) = y^t. \label{eq:targeted_attack}
\end{align} 

It was first observed in~\cite{szegedy2013intriguing} that deep neural networks are vulnerable to adversarial perturbation. Since then, researchers have further explored several attack strategies~\cite{AdvExp_Goodfellow, kurakin_adversarial_2017,Moosavi-Dezfooli_2016_CVPR,carlini_towards_2017,madry_towards_2019, fda} for generating adversarial examples. 
%including using the linear assumption behind a model (FGSM~\cite{AdvExp_Goodfellow} and DeepFool~\cite{Moosavi-Dezfooli_2016_CVPR}), 
%saliency maps [34], and evolutionary algorithms [29]. 
Fast gradient-based techniques like FGSM~\cite{AdvExp_Goodfellow} used the sign of gradient to generate a coarse approximation of optimal examples in one step.
% PGD~\cite{madry_towards_2019}, I-FGSM~\cite{kurakin_adversarial_2017}, FDA~\cite{fda} and C\&W~\cite{carlini_towards_2017}.
Further, I-FGSM~\cite{kurakin_adversarial_2017} extended the FGSM method by iterating the process with a smaller step size to improve the attack performance. Later, I-FGSM was further refined with multiple random starts in~\cite{madry2017towards} (PGD), or with momentum item in~\cite{dong_boosting_2018} (MI-FGSM). However, our experiments revealed almost negligible differences among I-FGSM, PGD, and MI-FGSM regarding the attack performance quantitatively measured by the proposed $\Delta$PSNR. Thus, we mainly use I-FGSM for comparative study in the paper. 
Carlini and Wagner~\cite{carlini_towards_2017} then introduced a more effective algorithm (referred to as CW) to best guarantee a successful attack with minimal perturbation injected. The CW method can be formulated under $l_{0}$, $l_{2}$, $l_{\infty}$ norms, and is able to take advantage of modern optimizers such as Adam~\cite{adam}   to generate  adversarial examples for the attack. %to measure the robustness of neural models. 

{\bf Defense.} \add{To deal with the threat caused by adversarial attacks, many defense techniques have also been proposed. Pre-processing based approaches like JPEG compression~\cite{dziugaite2016study}, random resizing~\cite{xie_mitigating_2018}, bit depth reduction~\cite{xu_feature_2017} and geometric self ensemble~\cite{lim_enhanced_2017} attempt to pre-process the input image to eliminate the effects caused by perturbations.
Another very effective way of defense is adversarial training that augments adversarial examples in training set~\cite{AdvExp_Goodfellow, Szegedy2014, 43405, kurakin_adversarial_2017, 2015Learning, alex2018} to directly improve the robustness of the underlying learned model for various vision tasks.} 

Existing attempts of adversarial attacks on VAE-based methods~\cite{Tabacof2016AdversarialIF, kos2018adversarial}  only focus on generative models like VAE-GAN~\cite{larsen2016autoencoding} and mainly use very low-resolution (e.g. 28$\times28$), simple digital number datasets like MNIST and SVHN~\cite{netzer2011reading}.
To the best of our knowledge, there has been no existing work on attacking full-resolution image compression frameworks. 
As aforementioned,  the stability of decoded reconstruction in image compression tasks is critical to vast applications.
Therefore, in this paper, we will first demonstrate how existing image compression models are affected by the untargeted attack and then show how to defend them for robust image compression. 

\textbf{Robustness Evaluation.} Most previous works have focused on attacking computer vision systems. Therefore, they generally use the performance drop in their corresponding tasks (e.g., classification accuracy in classification tasks) to evaluate the attack performance. For adversarial attacks in image-to-image tasks (super-resolution~\cite{choi_deep_2021, yin2018sr, choi2019evaluating}, optical flow~\cite{ranjan2019attacking}, denoising/deblurring~\cite{choi_deep_2021}, etc.), attack performance is often evaluated by qualitatively showing visual examples together with some objective metric results. To better evaluate the vulnerability of a given image compression model, a proper metric is highly desired to measure and compare the degree of vulnerability quantitatively. In this case, we propose the $\Delta$PSNR metric to quantify them effectively.

\section{Adversarial Examples}
\label{sec:method}

Images are mostly compressed for vast applications due to rate/bandwidth constraints in practice.
%for either content consumption or semantic understanding. Thus, 
The compression technology behind is mandated to guarantee interoperability across heterogeneous platforms, client devices, etc, which is often presented as an international standard or industrial recommendation that is reproducible and publicly accessible. 
Hence, we generally assume that the attacker can have full access to the NIC codec and model parameters to best generate adversarial examples for attacks~\cite{kos2018adversarial}. As a result, technically the proposed attack strategies belong to \textit{white-box} attack. Besides, such an adversarial attack is referred to as the ``untargeted'' attack without a designated purpose that particularly aims for misleading the semantic understanding of the reconstructed content. \add{In the companion supplemental material, we also demonstrate how a targeted attack can affect the outcome of NIC methods.}

Without making any modification to the parameters of pretrained encoder $f_E()$ and decoder $f_D()$ of a specific NIC method, the attacker wishes to inject perturbation $\bf n$ upon the original source image $\bf x$ to generate the adversarial example $\bf x^*= x + \mathbf{n}$. These images are respectively encoded and decoded to have the original reconstruction $\hat{\bf x} = f_D(f_E({\bf x}))$, and the adversarial $\hat{\bf x}^* = f_D(f_E({\bf x^*}))$. Both $\bf x^*$ and $\hat{\bf x}^*$ are constrained in the range of [0, 1].
%that has negligible perturbation to the original input $\bf x$, and mostly can not be differentiated visually.
%In this case, 
The generated ${\bf x}^*$ shall present negligible perturbation compared with the original input $\bf x$, while the decoded adversarial reconstruction $\hat{\bf x}^*$ is expected to largely differ from the original reconstruction $\hat{\bf x}$ with visible distortion. Therefore, the adversarial examples for such untargeted distortion attacks can be formally defined as follows:
\begin{equation}
\begin{aligned}
      % maximize \quad & \mathcal{D}(f_d{f_e(x)}, f_d{f_e(x+n)})\\
    maximize \quad & \mathcal{D}(\mathbf{\hat{x}, \hat{x}^*})\\
    s.t. \quad & \mathcal{D}(\mathbf{x, x^*}) < \epsilon,\\
    & \mathbf{x^*, \hat{x}^*} \in [0, 1].
    \label{eq:attack}  
\end{aligned}
\end{equation}
Here $\mathcal{D}$ is the distance metric that can be the MSE or MS-SSIM, and $\epsilon$ is the threshold used to control the input perturbation.

\add{{\bf Fast Threshold-constrained Distortion Attack (FTDA).}} To solve~\eqref{eq:attack}, we proposed the FTDA to optimize the adversarial example generation:
\begin{align}
    % \mathop{\arg\min}_{\mathbf{n}}{L_{d}} &= ||\mathbf{M}*\mathbf{n}||_2 + \lambda (1-||\hat{x}, \hat{x}^*||_2) 
    % \mathop{\arg\min}_{\mathbf{n}}{L_{d}} = ||\mathbf{n}||_2 + \lambda (1-||\hat{x}, \hat{x}^*||_2) \\
    % \label{eq:dloss}\\
    % \mathrm{sign}(||\mathbf{n}||_2^2 - \epsilon)||\mathbf{n}||_2^2 + (1-\mathrm{sign}(||\mathbf{n}||_2^2 - \epsilon))(1-||\hat{x}, \hat{x}^*||_2^2) \\
    \mathop{\arg\min}_{\mathbf{n}}{L_{d}} =\left\{ \begin{array}{lc}
       \frac{||\mathbf{n}||_2^2}{\mathbf{N}}, & \frac{||\mathbf{n}||_2^2}{\mathbf{N}} \geq \epsilon, \\
        1-\frac{||\hat{\bf x}-\hat{\bf x}^*||_2^2}{\mathbf{N}}, & \frac{||\mathbf{n}||_2^2}{\mathbf{N}} < \epsilon.
    \end{array}
    \right.
    \label{eq:dloss}
\end{align}
where $\epsilon$ is the $l_2$ threshold and N is the total num of image pixels. In this paper, we mainly use $l_2$ distance for $\mathcal{D}$ because  $l_2$ related PSNR metric is widely used for image compression quality measurement. \add{We also show some examples in Fig.~\ref{fig:metrics_attack} in the supplemental material that other distance metrics, such as the $l_1$ distance and  MS-SSIM~\cite{MS-SSIM}, can be used to guide the example generation.}

Equation~\eqref{eq:dloss} is a piecewise function. When augmented noise is strong, e.g.,   $\frac{||\mathbf{n}||_2^2}{\textbf{N}} \geq \epsilon$, the objective function is first to decrease the noise intensity until $\frac{||\mathbf{n}||_2^2}{\textbf{N}} < \epsilon$; 
hereafter,  we use the negative $l_2$ distance between the original reconstruction $\hat{\bf x}$ and adversarial reconstruction $\hat{\bf x}^*$ to maximize their distance. The threshold $\epsilon$ can enforce the same level of perturbation added in the input and maximize the output distortion in the meantime. We can easily adapt $\epsilon$ to get proper adversarial examples as expected.  %\textcolor{red}{Tong Tong, is different distance correlated with loss function used in optimizing the NIC?}
The proposed perturbation generation strategy in \eqref{eq:dloss} is a general approach that is effective for any NIC models used in practice. 
As will be shown in Sec.~\ref{sec:attack_defense}, these generated adversarial examples can be used to retrain existing NIC
 models and effectively mitigate the artifacts induced by adversarial attacks and successive image compression.

Besides, existing attack approaches can also be extended to solve \eqref{eq:attack}, where certain adjustments are implemented to adapt them for the untargeted distortion attack. Here we choose commonly used I-FGSM~\cite{kurakin_adversarial_2017} and CW-$l_2$~\cite{carlini_towards_2017} for evaluation in the following section. 

\textbf{I-FGSM.}
Here is the optimization function used for I-FGSM. $\operatorname{sgn}()$ is a sign function applied to the backpropagated gradient. At the $i$-th step,
\begin{equation}
\mathbf{x}^*_{(i+1)}=\mathbf{x}^*_{(i)}+\frac{\epsilon}{T} \operatorname{sgn}\left(\nabla\left\|\hat{\mathbf{x}}_{(i)}^*-\hat{\mathbf{x}}\right\|_{2}\right).
\end{equation}
Similar to~\eqref{eq:dloss}, here $\epsilon$ also controls the level of input perturbation. $T$ is the iterations needed for example generation which is set to 16 in default. 
%\revisea{We also include PGD~\cite{madry_towards_2019} with multi random starts and MI-FGSM (Momentum I-FGSM)~\cite{dong_boosting_2018} for comparison.}%The default iteration number T is set to 16.

\textbf{CW-$l_2$.}
The original version of the CW attack was designed for the targeted attack as:
\begin{equation}
\begin{aligned}
    minimize & \quad ||\delta||_p+c \cdot f(x+\delta),\\ 
    such\;that & \quad x+\delta \in[0,1].
\end{aligned}
\label{eq:cw}
\end{equation}
where $||\delta||_p$ is the $l_p$ norm of perturbation $\delta$ and $f(x+\delta)$ is predefined objective function. For a fair comparison, the $l_2$ norm is used in our experiments and the negative $l_2$ distance is applied as the objective function. The final adjusted CW-$l_2$ used in this work is formulated as:
\begin{equation}
\begin{aligned}
    minimize & \quad \frac{||\mathbf{n}||_2^2}{\mathbf{N}}+ \left(1-c \cdot \frac{||\mathbf{\hat{x}}^*-\mathbf{x}||_2^2}{\mathbf{N}}\right),\\ 
    such\;that & \quad \mathbf{\hat{x}^*, x^*} \in[0,1].
\end{aligned}
\label{eq:cw-ours}
\end{equation}

Here $c$ is a constant that needs to be suitably chosen. In order to achieve the optimal attack performance and also control the input perturbation level, we use a binary search strategy to find the optimal $c$ following the implementation suggested by the authors of the original CW method~\cite{carlini_towards_2017}. For each  $c$ under consideration, we run 1000 iterations in default to minimize~\eqref{eq:cw-ours} with the Adam optimizer.

\section{Attack Evaluation}
\label{sec:exp}
\subsection{$\Delta$PSNR}

% \begin{figure}[htbp]
% \centering
% \includegraphics[width=0.25\textwidth]{./figs/dpsnr_linear.pdf}%
% \includegraphics[width=0.25\textwidth]{./figs/metrics_with_clip.pdf}
% \includegraphics[width=0.25\textwidth]{./figs/metrics_without_clip.pdf}
% \caption{\add{{\bf Vulnerability metric comparison.} $\Delta$PSNR can represent the quality degradation more \textcolor{red}{constantly} at various input perturbation levels controlled by $\epsilon$ in~\eqref{eq:dloss}. A pretrained Ball\'e2018 model and Kodak images are used in this example.}}
% \label{fig:metrics_vi}
% \end{figure}

\begin{figure*}[htbp]
\centering
\begin{subfigure}[b]{0.33\textwidth}
    \includegraphics[scale=0.445]{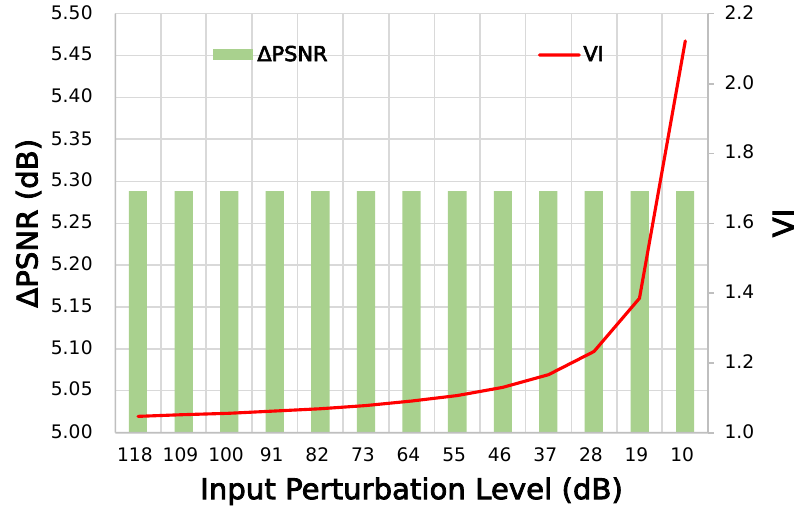}
    \caption{}
    \label{subfig:linear}
\end{subfigure}%
\begin{subfigure}[b]{0.33\textwidth}
    \includegraphics[scale=0.45]{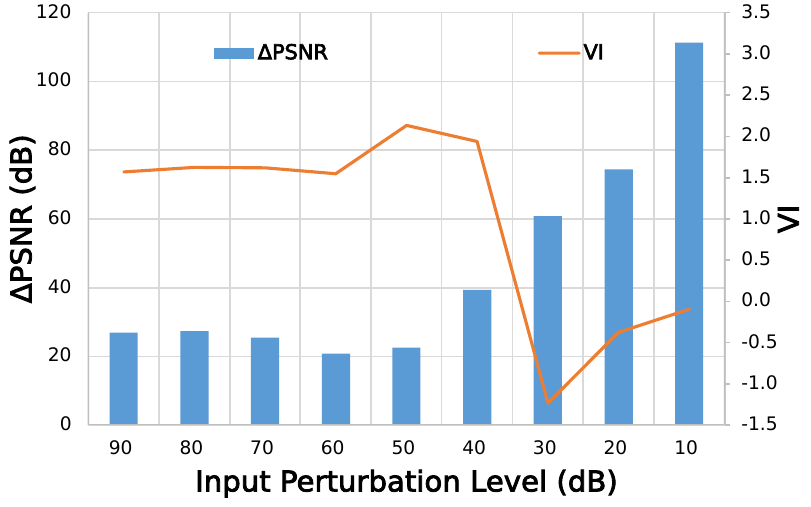}
    \caption{}
    \label{subfig:dpsnr_nic}
\end{subfigure}
\begin{subfigure}[b]{0.33\textwidth}
    \includegraphics[scale=0.445]{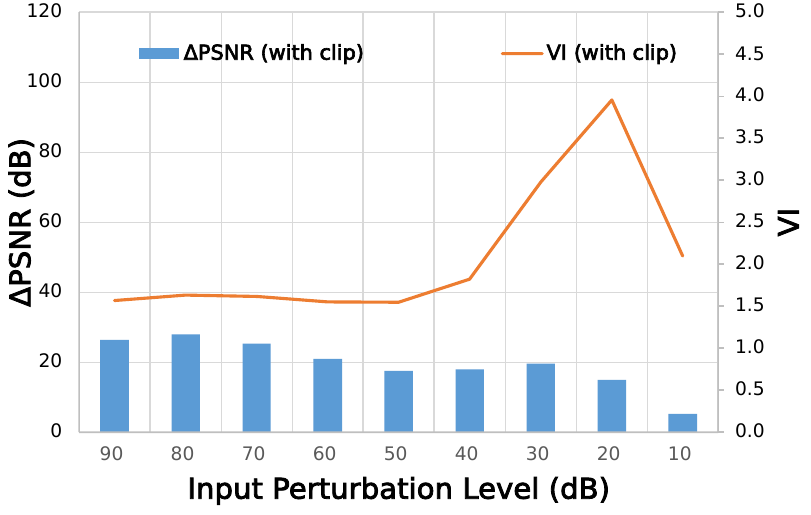}
    \caption{}
    \label{subfig:dpsnr_clip}
\end{subfigure}
\caption{Vulnerability metric comparison. $\Delta$PSNR can better represent the quality degradation at various input perturbation levels controlled by $\epsilon$ in~\eqref{eq:dloss}. Kodak images are used in this example. (a) A toy linear network using one-layer convolution for both encoder and decoder; (b) non-linear NIC; (c) NIC with clipping to ensure $\mathbf{x^*, \hat{x}^*} \in [0, 1]$ as in \eqref{eq:attack}. For (b) and (c), A pretrained Ball\'e2018 model is tested.}
\label{fig:metrics_vi}
\end{figure*}

Recalling the attack formulation in \eqref{eq:attack}, we hope the negligible perturbation at the input would bring severe distortion to the output, which implies that the underlying network enlarges the input perturbation through layer-wise computation. In this case, we suggest measuring the vulnerability by the level of the perturbation amplification. Therefore, a $\Delta$PSNR is used:
\begin{align}
\Delta&\text{PSNR}  = \nonumber\\
& -10\log_{10}{\frac{||\mathbf{{x}^*-{x}}||_2^2}{||\mathbf{\hat{x}^*-\hat{x}}||_2^2}} = \text{PSNR}_\text{in} - \text{PSNR}_\text{out}.
\label{eq:vi}
\end{align} 
%where the ratio is calculated by $l_2$ (mse) distance first and then convert to dB format as in~\eqref{eq:vi}.
$\Delta\text{PSNR}$ means the quality degradation measured by PSNR. As seen, $\Delta\text{PSNR}>0$ means that the input perturbation is magnified by the network so the output distortion is larger than the input perturbation, while $\Delta\text{PSNR}<0$ means that the input perturbation is suppressed to a smaller level.

Alternatively, a Vulnerability Index (VI) metric was used in~\cite{choi_deep_2021} for evaluating the vulnerability of image processing networks. It is calculated as the PSNR ratio between the input and output:
\begin{align}
\text{VI} = \frac{-10\log_{10}||{\bf x^*} - {\bf x}||_2^2}{-10\log_{10}||\hat{\bf x}^* - \hat{\bf x}||_2^2} = \text{PSNR}_\text{in}/\text{PSNR}_\text{out}.
\label{eq:vi_old}
\end{align}

%Larger VI values indicate that the tested model has a higher vulnerability. 

Figure~\ref{fig:metrics_vi} provides a comparison of these two measurements. For a tested model with fixed parameters, the VI metric shows obvious sensitivity to the absolute level of input perturbation even for linear networks as in Fig.~\ref{subfig:linear}. However, the magnification level presented by $\Delta$PSNR is more stable across different perturbation levels.
% we can generally assume the constant scaling factor for input perturbation magnification. 
As a result, the $\Delta$PSNR is a preferred metric to measure the intrinsic vulnerability of an image compression model without being biased by the absolute magnitude of input perturbation. More details are given in the supplemental material.

In the following sections, $\Delta$PSNR will be used for performance evaluation. For measuring the attack performance, higher $\Delta$PSNR means better attack performance; while for the defense efficiency, lower $\Delta$PSNR indicates better defense efficiency.

% This section follows the optimization strategies proposed in \eqref{eq:dloss} to measure the robustness of the popular NIC methods. Extensive simulations reveal that distortion attack works effectively and leads to visible impairments in adversarial reconstruction \edit{regardless of underlying NIC methods, quality scales and optimization metrics, etc.}
%\delete{Since the network parameters are fixed for all the following experiments, the simulated quantization $\mathbf{Q}$ mentioned in Fig.~\ref{fig:vae} with uniform noise can be skipped during the generation stage of adversarial examples to achieve more stable gradient backward propagation. Note that the quantization step is always retained during inference.}

% \begin{figure*}[t]
% \centering
% \includegraphics[scale=0.2]{./figs/distortion.pdf} 
% \caption{Distortion attack on Kodak dataset. ($1^{st}$ row) adversarial examples; ($2^{nd}$ row) its corresponding reconstructions.}
% \label{kodak}
% \end{figure*}

\subsection{Channel-wise Activation Variation (CAV)}
\begin{figure}[htbp]
\centering
\includegraphics[width=0.36\textwidth]{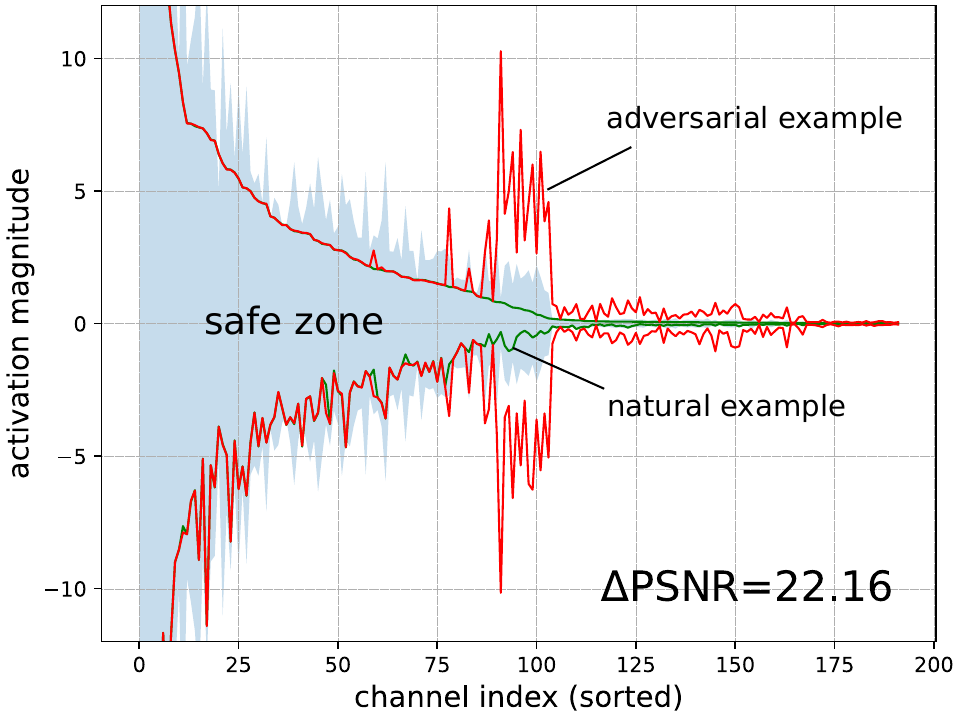}
 \caption{{\bf Channel-wise Activation Variation (CAV).} \add{The range of channel-wise activation magnitude at the bottleneck layer for the original natural image and its corresponding adversarial example is illustrated. The channels are sorted in descending order according to the range of activation magnitude of the original natural image. An MS-SSIM optimized Ball\'e2018 model is exemplified.}} 
\label{fig:activation_demo}
\end{figure}

Besides, we offer another perspective by visualizing and comparing the channel-wise activation in latent feature space. As shown in Fig.~\ref{fig:activation_demo}, the \textbf{safe zone} indicates the range of activation magnitudes counted over 10,000 training images. The activation intensity of the original natural image (without noise injection) is very likely to stay within this region. However, for the activation intensity of adversarial examples, noticeable outliers are observed, especially for some channels that are much less activated by natural images. Such qualitative visualization of CAV  well supplements the objective $\Delta$PSNR. Having them together provides a comprehensive evaluation to study the impact of the adversarial attack on image compression models.

\subsection{Adversarial Example Generation Settings}
\add{In order to find optimal settings for adversarial example generation, a variety of experiments are conducted first.
% Previous sections report that existing NIC models are vulnerable to adversarial attacks regardless of the implementation methods, loss functions, and compressed bitrates utilized. This section further shows that adversarial attack works in general when applying different noise thresholds and distance measurements in~\eqref{eq:dloss} to inject perturbation.
}
\subsubsection{Attack Strategy Comparison}
\begin{figure}[t]
\centering
\includegraphics[width=0.48\textwidth]{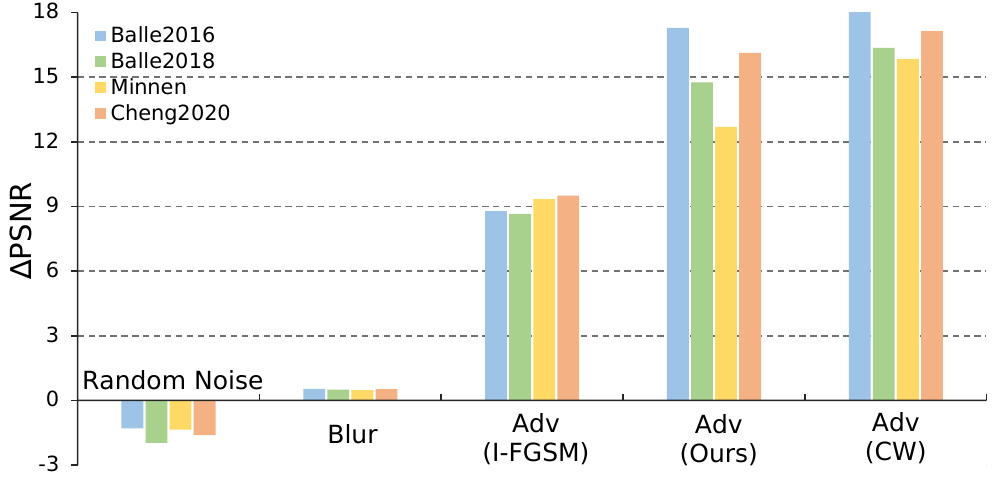}
\caption{{\bf Attack performance of different strategies measured by $\Delta \text{PSNR}$.} The input perturbation is controlled at the same level of 1e-4 for different methods including the gaussian random noise and blur. Results for each method are averaged over all quality scales and metrics in Table~\ref{tab:quality} on Kodak images.}
\label{fig:vi}
\end{figure}

 Figure~\ref{fig:vi} presents the attack performance of I-FGSM, CW-$l_2$, and the proposed FTDA (Ours). 
As a comparative anchor, we also provide the $\text{PSNR}$ degradation caused by random noise attack and blurry degradation. As you can observe, compression networks tend to suppress the effect of random noise but dramatically amplify the adversarial perturbation generated by other attack methods. Among them, the CW-$l_2$ algorithm shows slightly better performance than our proposed method measuring by $\Delta$PSNR.

\begin{figure}[htbp]
\centering
\includegraphics[width=0.49\textwidth]{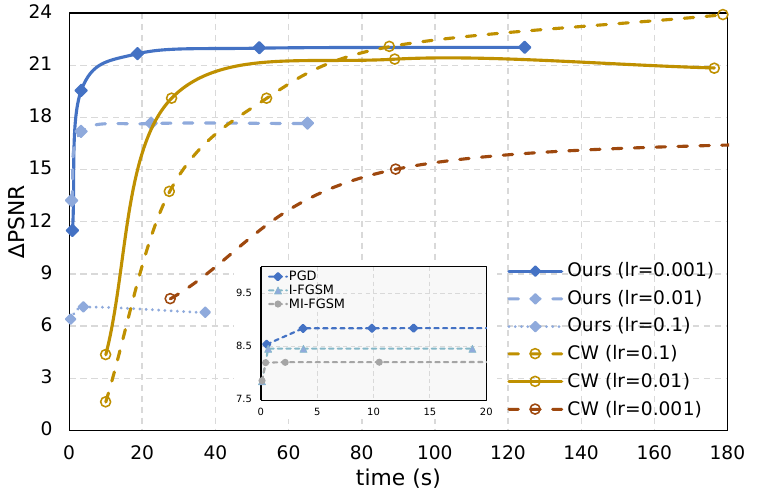}
\caption{{\bf Attack performance vs. complexity for different attack strategies.} A pretrained Ball\'e2018 model is used here as an example. 
%Results for each method's default setting are marked with red circles. The CW-$l_2$ strategy takes more than 10$\times$ longer duration to reach a similar performance as our proposed strategy in~\eqref{eq:dloss}.
Various learning rates at 0.001, 0.01, and 0.1 are tested for CW and the proposed method. For each learning rate, binary search steps are modified accordingly to obtain the CW performance at different time complexities. Our proposed method performs better when having complexity limitations, while the CW can achieve better performance with much higher computational complexity. Three FGSM-based methods (I-FGSM, MI-FGSM and PGD) show very close $\Delta$PSNR (e.g., between 8 and 9).}
\label{fig:complexity}
\end{figure}

In addition to the attack performance, computational complexity is another vital factor for the successful adoption of a potential attack strategy. Therefore,
when comparing different attack methods, we also take the computational complexity (measured by the time duration in seconds for example generation) into consideration. 
%by modifying the iteration numbers for each method. 

The results are illustrated  in Fig.~\ref{fig:complexity}. As mentioned earlier, I-FGSM is very fast because of its gradient sign computation. But the coarse adversarial examples it generates are less effective with the least $\Delta$PSNR (see the subplot embedded). The CW-$l_2$ in~\eqref{eq:cw-ours} needs to search for the best $c$ in order to get optimal adversarial attack performance. The searching process  inevitably leads to much higher computational complexity. If setting no computational complexity constraint for all strategies, the CW-$l_2$ can reach at the best attack performance. 
However, 
it is impractical to use the CW-$l_2$ for large-scale adversarial example generation and subsequent adversarial training. In this case, our proposed FTDA finds a justified balance between complexity and performance. 

%still work when having different noise injection optimization strategies in \eqref{eq:dloss}.
\begin{figure}[t]
\centering
\includegraphics[scale=0.15]{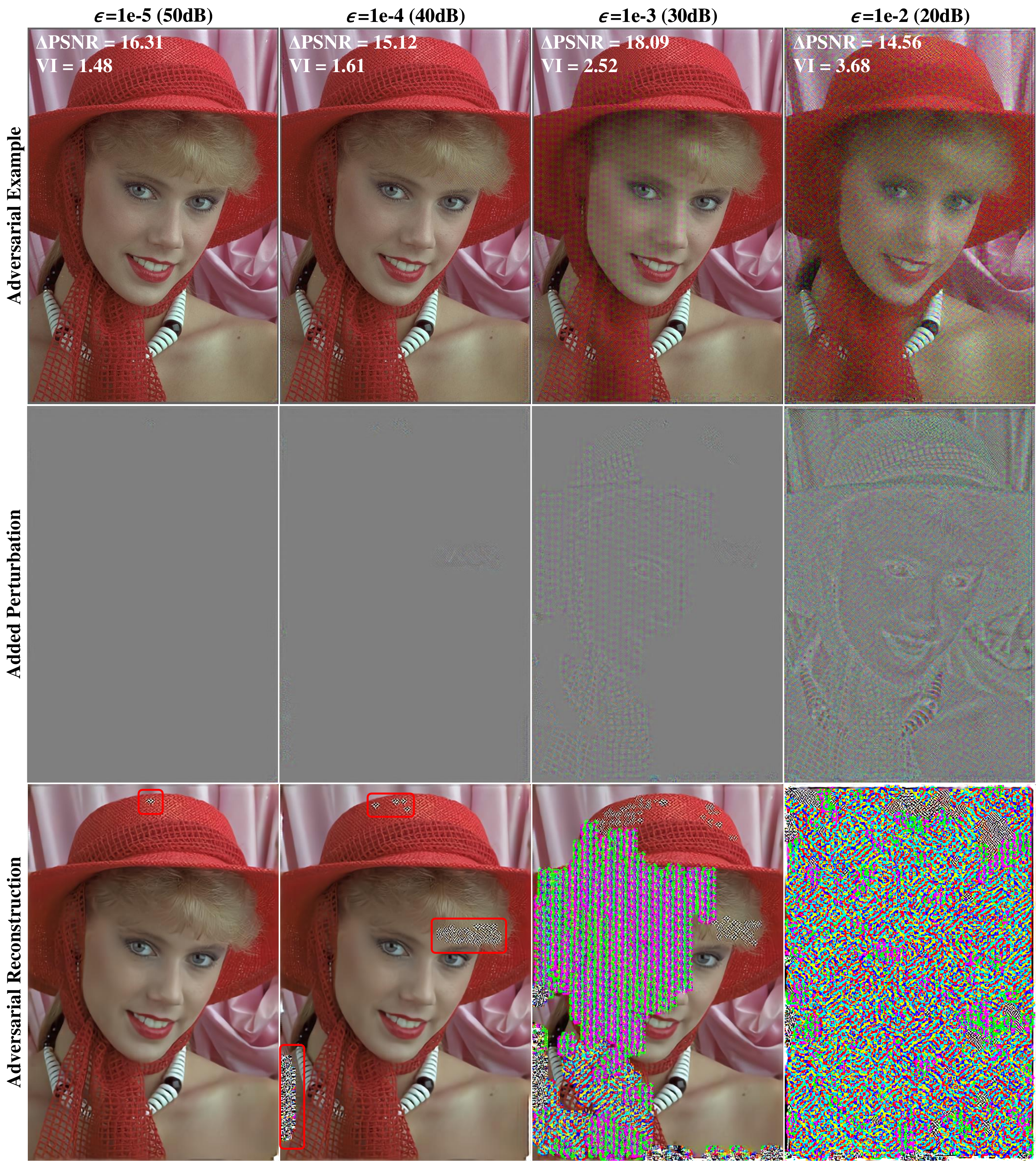}
\caption{{\bf Impact of input perturbation level $\epsilon$.} Illustration of adversarial examples at the first row, associated perturbation at the second row, and corresponding decoded reconstructions at the third row for different perturbation levels. $\epsilon$ is adjusted to adapt the intensity of input perturbation, i.e., larger $\epsilon$ indicates stronger perturbation. Both $\Delta$PSNR and VI are presented. Still here VI always tends to increase with the absolute perturbation level.}
\label{fig:epsilon}
\end{figure}
\subsubsection{Perturbation Level $\epsilon$}
Figure~\ref{fig:epsilon} further visualizes adversarial examples when having different levels of input perturbation adapted by the  $\epsilon$. The results are consistent with the intuition, i.e., larger $\epsilon$ leads to stronger noise for the input image, and thus yields more severely-degraded reconstructions. Surprisingly, even with negligible visual artifacts when  $\epsilon$ = 1\text{e-}5, impaired pixels are still presented in decoded reconstruction, implying the general vulnerability of NIC models. As also observed in Fig.~\ref{fig:epsilon}, the input perturbation would become visible when $\epsilon$ is increased to 1e-3. To control the input perturbation basically imperceptible, the default noise threshold $\epsilon$ in \eqref{eq:dloss} is set to 1e-4 for the following experiments, having the PSNR between the original natural image and corresponding adversarial example equal to 40 dB.%With the perturbation getting bigger, the reconstruction will finally be submerged by distortion.
%  More results can be found in Appendix~\ref{apdx:visual}.

\subsection{Evaluation on NICs}
\label{sec:method_distorion}
% Previous sections report that existing NIC models are vulnerable to adversarial attacks regardless of the implementation methods, loss functions, and compressed bitrates utilized. This section further shows that adversarial attack works in general when applying different noise thresholds and distance measurements in~\eqref{eq:dloss} to inject perturbation.
Experiments in the previous section have shown some preliminary evidence that existing NIC models are vulnerable to adversarial attacks. In this section, we will perform more tests and further explore the influence of different NIC settings on the model vulnerability.
\subsubsection{General Observation} 
We choose a variety of VAE-based NICs listed in Table~\ref{tab:methods} to demonstrate the general existence of vulnerability issue among these recently emerged solutions. These NICs exemplify major milestones during the development of learning-based image coding.
%For any image compression codec, the distortion is always a fundamental issue and should raise more concerns.

%improving the general architecture in Fig.~\ref{fig:vae} for better performance. 
Theoretically, image coding efficiency is highly related to the efficiency of nonlinear transform and entropy context modeling. Earlier attempts made in Ball\'e2016~\cite{balle2016end} first introduced the GDN (Generalized Divisive Normalization) with convolutional layers and applied a simple factorized entropy model that was further improved in Ball\'e2018~\cite{balle2018variational} and Minnen~\cite{minnen2018joint} by hyperprior, and joint hyperprior and autoregressive neighbors for better entropy modeling; Because of the superior efficiency by using joint hyper prior and autoregressive neighbors, succeeding works almost reused the same mechanism in context modeling.

Given that convolution is calculated using equal weights at all spatial locations, {Cheng2020}~\cite{Cheng2020Learned} and NLAIC~\cite{chen2020} suggested the attention mechanism to aggregate local or nonlocal correlations to enhance the performance. Previous discussions mainly applied the MSE or MS-SSIM as the loss function in training while HiFiC~\cite{mentzer2020high} combined the MSE, LPIPS (Learned Perceptual Image Patch Similarity), and GAN (Generative Adversarial Network) loss to noticeably improve the perceptual quality of reconstructed images.

In addition to the coding efficiency, other aspects were considered as well. For example, the aforementioned models with floating point computation were platform-dependent and complexity intensive. In Weixin2021~\cite{weixin2021fix}, it developed the fixed point strategy to optimize the NLAIC~\cite{chen2020}, yielding negligible performance loss but significant complexity reduction. Note that nonlinear transforms in~\cite{balle2016end,balle2018variational,minnen2018joint,chen2020,Cheng2020Learned} were not invertible and could not avoid the information loss, the InvCompress~\cite{xie2021enhanced} used invertible neural network (INN) as its core transform to resolve this issue.

We generally utilize open-source code and pretrained models of these methods (e.g., compressAI\footnote{https://github.com/InterDigitalInc/CompressAI}~\cite{begaint2020compressai}, NLAIC\footnote{https://njuvision.github.io/NIC/}, Weixin2021\footnote{https://njuvision.github.io/fixed-point/}, HiFiC\footnote{https://github.com/tensorflow/compression/tree/master/models/hific} and InvCompress\footnote{https://github.com/xyq7/InvCompress}) to avoid any ambiguity and inconsistency.
For those missing models, 
% \delete{such as MS-SSIM optimized Minnen~\cite{minnen2018joint} and Cheng 2020~\cite{Cheng2020Learned},} 
we strictly follow the training procedures to reproduce them from scratch.
%As seen, these methods cover most popular components (GDN, auto-regressive context model, attention machanism, etc.), quality metrics and state-of-the-art performance. Note that the implementation of these methods are based their on open-sourced code (compressAI\footnote{https://github.com/InterDigitalInc/CompressAI}~\cite{begaint2020compressai}, NLAIC\footnote{https://njuvision.github.io/NIC/}, Weixin 2021\footnote{https://njuvision.github.io/fixed-point/}, HiFiC\footnote{https://github.com/tensorflow/compression/tree/master/models/hific} and InvCompress\footnote{https://github.com/xyq7/InvCompress}). Most pretrained models are also provided by the above projects except for MS-SSIM optimized models for Minnen~\cite{minnen2018joint} and Cheng 2020~\cite{Cheng2020Learned}, which are not provided so we carefully trained these models basing on the open-sourced code. Kodak~\cite{kodak} images are used for evaluating the distortion attacking result.
The hyperparameter $\lambda$s used for different quality scales follow the setting provided by CompressAI as listed in Table~\ref{tab:quality}.
\begin{figure}[t]
\centering
\includegraphics[width=0.5\textwidth]{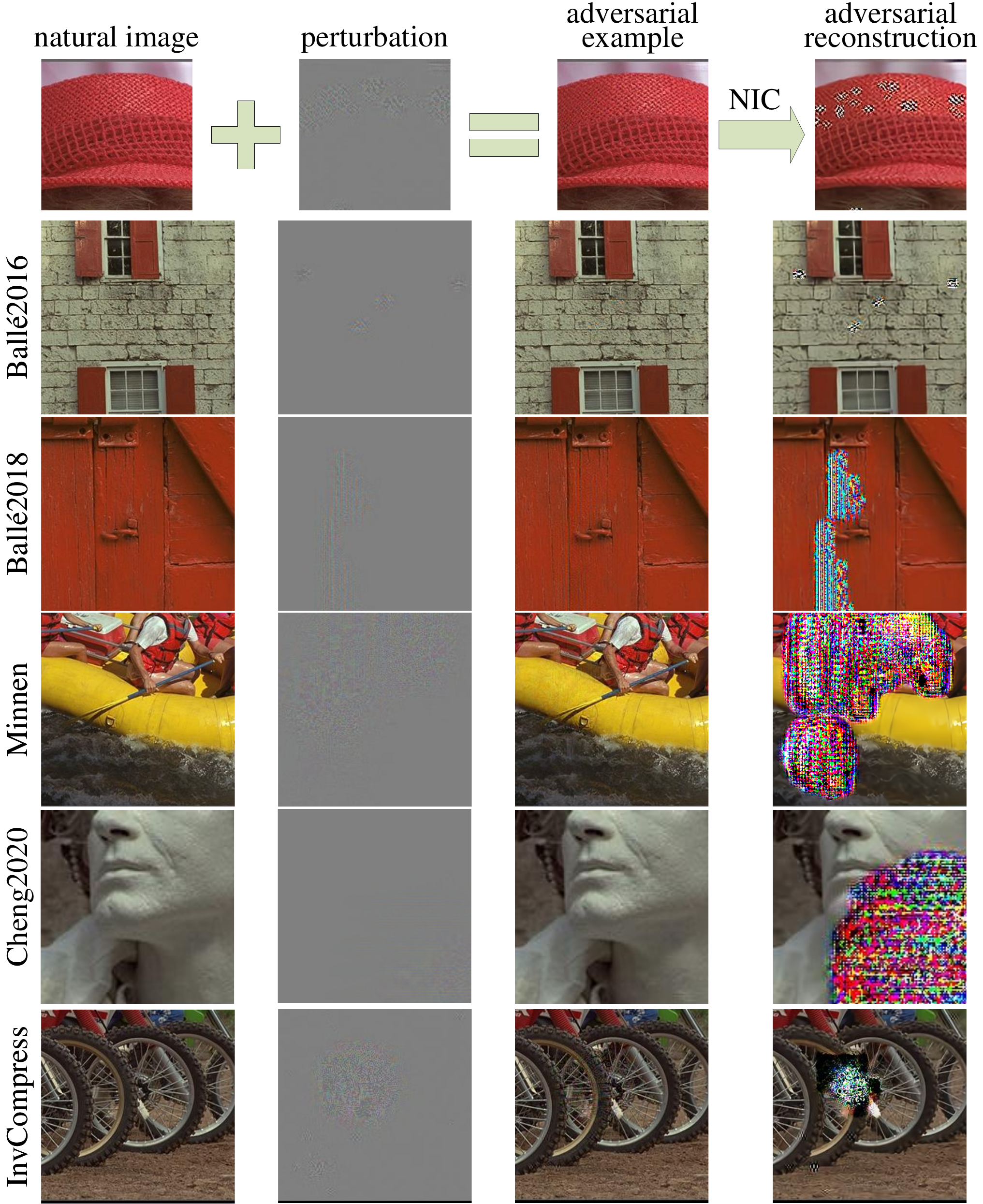} 
\caption{\add{{\bf Examples of Model Vulnerability Incurred Reconstruction Distortions.} 
More quantitative comparisons between different methods can be found in Table~\ref{tab:nics}.}}
\label{fig:example}
\end{figure}
For a fair comparison, we run 1k steps with a learning rate at 0.01 with three times learning rate decay and noise threshold $\epsilon$ at 1e-4 to generate adversarial examples for each NIC method listed in Table~\ref{tab:methods} using images from Kodak dataset~\cite{kodak}. Adam~\cite{adam} optimizer is used to optimize the \eqref{eq:dloss}.

 In Fig.~\ref{fig:example}, the adversarial inputs and corresponding reconstructions are visualized for multiple methods\footnote{We use different images for diverse NIC solutions to exemplify the severe quality degradation of decoded images incurred by the model vulnerability. Such visible distortions are presented for other images as well.}. Visible impairments are clearly presented in these reconstructions.
Quantitative results in Table~\ref{tab:nics} with significant PSNR degradation (e.g., $> 10$ dB) also confirm the subjective loss, 
 implying the general vulnerability issue in prevalent learning-based image coding frameworks regardless of their transforms, entropy models, and optimization methods utilized (see more details in Table~\ref{tab:methods}).
 % (even for methods using invertible neural network~\cite{xie2021enhanced} and \edit{fixed-point computation~\cite{weixin2021fix}}).} 
% \edit{As seen in Table~\ref{tab:nics}, these methods show clear vulnerability to adversarial attacks regardless of different entropy approximations. 
Even for InvCompress~\cite{xie2021enhanced} that applied a quite different invertible neural network to avoid information loss, the model vulnerability is still presented with obvious distortion in \edit{Fig.~\ref{fig:example}}, which might be caused by the use of (non-linear) feature enhancement modules before and after the invertible neural network transforms.

% \begin{table}[htbp]
% \centering
% \caption{\textcolor{blue}{The bitrate increase ($\Delta$bpp), $\Delta$PSNR and $\Delta$MS-SSIM of adversarial examples ($\epsilon$=1e-4) compared with original images averaged on Kodak dataset across all 6 quality levels (except for HiFiC, which only has 3 available pretrained levels). The ``mse'' and ``ms-ssim'' indicate the MSE and MS-SSIM optimized model respectively.}}
% \label{tab:nics}
% \setlength{\tabcolsep}{4pt}
% \begin{tabular}{l|c|c|c} 
% \hline
% \textbf{Methods} & \textbf{Metrics} & \textbf{$\Delta$\textbf{bpp}} & $\Delta$\textbf{PSNR}\\ 
% \hline
% \multirow{2}{*}{\textbf{Ball\'e2016}} & mse & +3.69\%                    & 14.03\\
%                                      & ms-ssim             & +5.78\%                    & {20.55}\\ 
% \hline
% \multirow{2}{*}{\textbf{Ball\'e2018}} & mse & +7.94\% & 11.38\\
%                                      & ms-ssim  & {+8.80\%}  & {16.45}\\ 
% \hline
% \multirow{2}{*}{\textbf{Minnen}}    & mse & +9.49\% & 10.39\\
%                                     & ms-ssim & +11.42\% & 14.96\\ 
% \hline
% \multirow{2}{*}{\textbf{Cheng2020}} & mse & {+113.49\%} & {18.27}\\
%                                     & ms-ssim       & +12.30\% & 13.98\\ 
% \hline
% \textbf{InvCompress}                         & mse & +6.50\% & 13.03\\
% \hline
% \multirow{2}{*}{\textbf{NLAIC}} & mse & +17.79\% & 10.22\\
% & ms-ssim & +8.17\% & 15.00\\
% \hline
% \textbf{Weixin2021} & mse & +4.56\% & 11.54\\
% \hline
% \textbf{HiFiC} & LPIPS & +4.08\% & 13.51\\
% \hline
% \end{tabular}
% \end{table}

\begin{table}[htbp]
\centering
\caption{\add{The bitrate increase ($\Delta$bpp) and $\Delta$PSNR  of adversarial examples ($\epsilon$=1e-4) compared with original images averaged on Kodak dataset across all 6 quality levels (except for HiFiC, which only has 3 available pretrained levels).} The ``mse'' and ``ms-ssim'' indicate the MSE and MS-SSIM optimized model respectively.}
\label{tab:nics}
\setlength{\tabcolsep}{4pt}
\begin{tabular}{l|c|c|c|c|c} 
\hline
\textbf{Methods} & \textbf{Metrics}  & \textbf{bpp (ori)} & \textbf{bpp (adv)} & \textbf{$\Delta$\textbf{bpp}}          & $\Delta$\textbf{PSNR}\\ 
\hline
\multirow{2}{*}{\textbf{Ball\'e2016}} & mse     & 0.4415              & 0.4578              & +3.69\%                    & 14.03                     \\
                                     & ms-ssim & 0.3443              & 0.3642              & +5.78\%                    & {20.55}    \\ 
\hline
\multirow{2}{*}{\textbf{Ball\'e2018}} & mse     & 0.4569              & 0.5044              & +10.40\%                   & 11.38                     \\
                                     & ms-ssim & 0.3301              & 0.3686              & {+8.80\%}  & {16.45}    \\ 
\hline
\multirow{2}{*}{\textbf{Minnen}}    & mse     & 0.4234              & 0.4636              & +9.49\%                    & 10.39                     \\
                                     & ms-ssim & 0.3136              & 0.3494              & +11.42\%                   & 14.96                     \\ 
\hline
\multirow{2}{*}{\textbf{Cheng2020}} & mse & 0.3966 & 0.8467 & {+113.49\%} & {18.27}\\
                                    & ms-ssim & 0.3050              & 0.3425              & +12.30\%                   & 13.98                     \\ 
\hline
\textbf{InvCompress}                         & mse     & 0.2967              & 0.3160              & +6.50\% & 13.03 \\
\hline
\multirow{2}{*}{\textbf{NLAIC}} & mse & 0.4144 & 0.4748 & +17.79\% & 10.22\\
& ms-ssim & 0.3348 & 0.3622 & +8.17\% & 15.00\\
\hline
\textbf{Weixin2021} & mse & 0.3174 & 0.3319 & +4.56\% & 11.54\\
\hline
\textbf{HiFiC} & LPIPS & 0.2574 & 0.2679 & +4.08\% & 13.51\\
\hline
\end{tabular}
\end{table}
%More numerical comparison results can be found in Table~\ref{tab:nics} that lists the reconstruction quality of the original no noise input and of adversarial input averaged on 24 Kodak images. 
% \edit{Note that all PSNR and MS-SSIM values are measured against the uncompressed, original image $\bf x$ for quantitative evaluation in this study.} 

\begin{figure*}[t]
    \centering     \subcaptionbox{Baseline\label{fig:entropy_baseline}}{\includegraphics[scale=0.98]{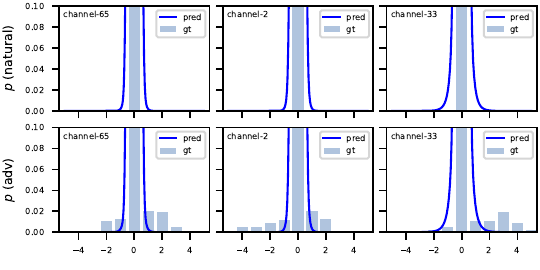}}
    \subcaptionbox{AT\label{fig:entropy_at}}{\includegraphics[scale=0.98]{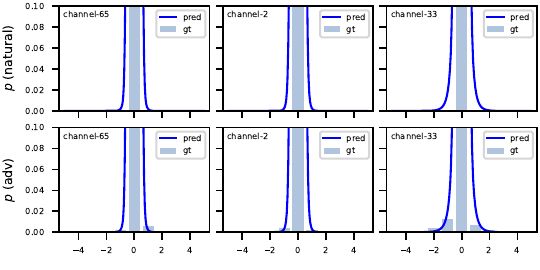}}     \caption{\add{Illustration of the divergence between predicted distribution (pred) by the entropy estimator and actual distribution (gt) of the latent features to be entropy coded. The model used here is MS-SSIM optimized Ball\'e2018 with quality level q=5. `Kodim03' in Kodak is used as the natural image and for generating the corresponding adversarial example. Three channels with the largest divergence are exemplified for illustration.}}
    \label{fig:entropy}
\end{figure*}

\begin{figure*}[t]
\centering
\subcaptionbox{Baseline (MS-SSIM opt.)\label{subfig:activation_a}}{\includegraphics[width=0.24\textwidth]{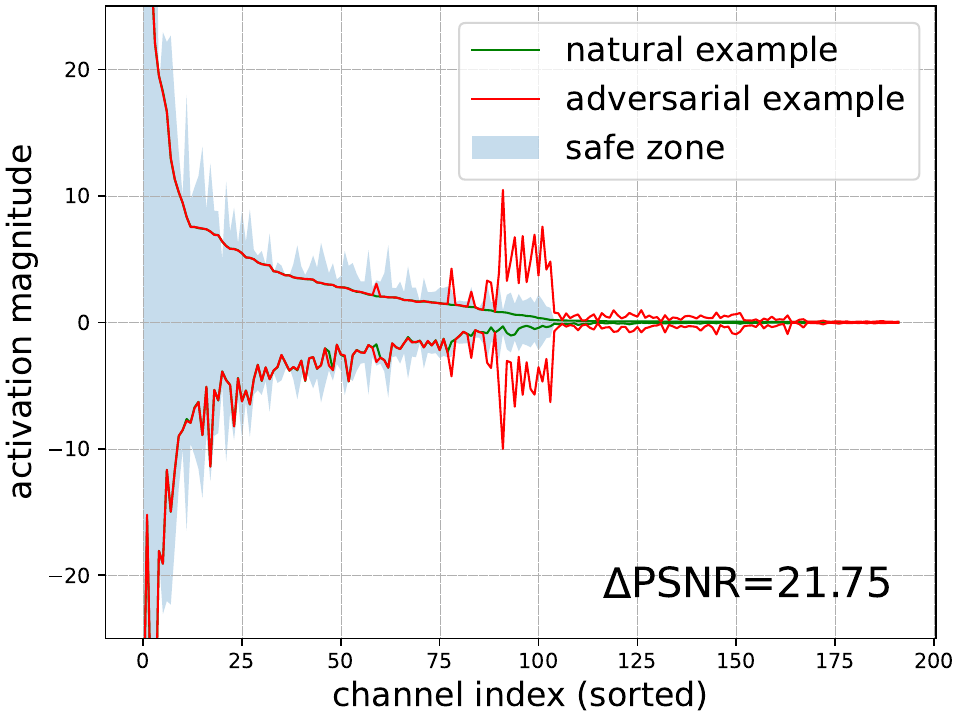}}
\subcaptionbox{AT (MS-SSIM opt.)\label{subfig:activation_b}}{\includegraphics[width=0.24\textwidth]{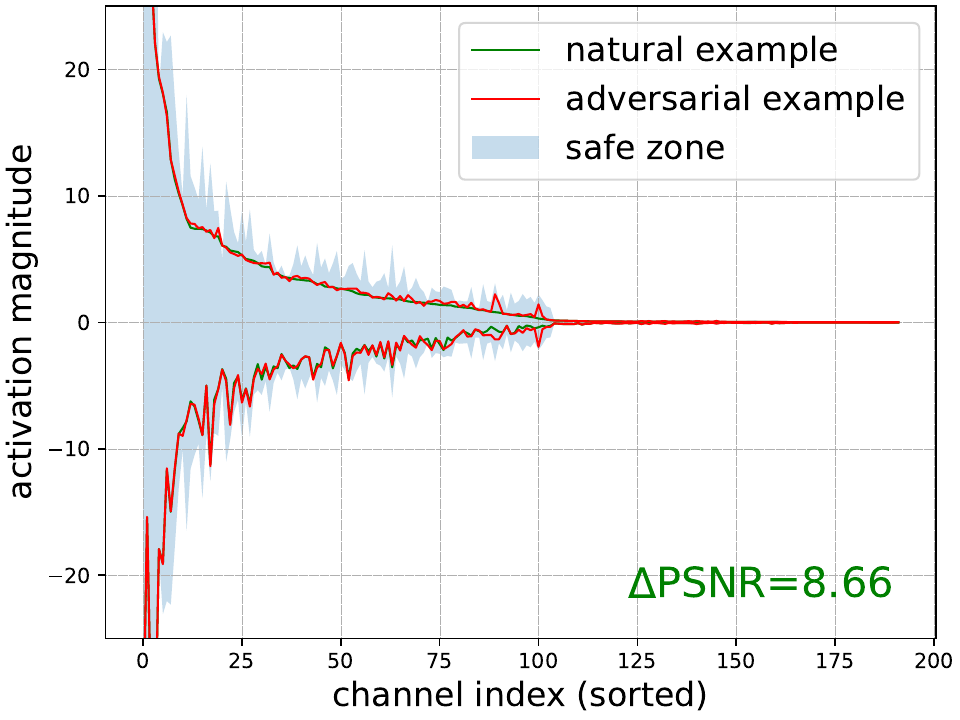}}
\subcaptionbox{Baseline (MSE opt.)\label{subfig:activation_c}}{\includegraphics[width=0.24\textwidth]{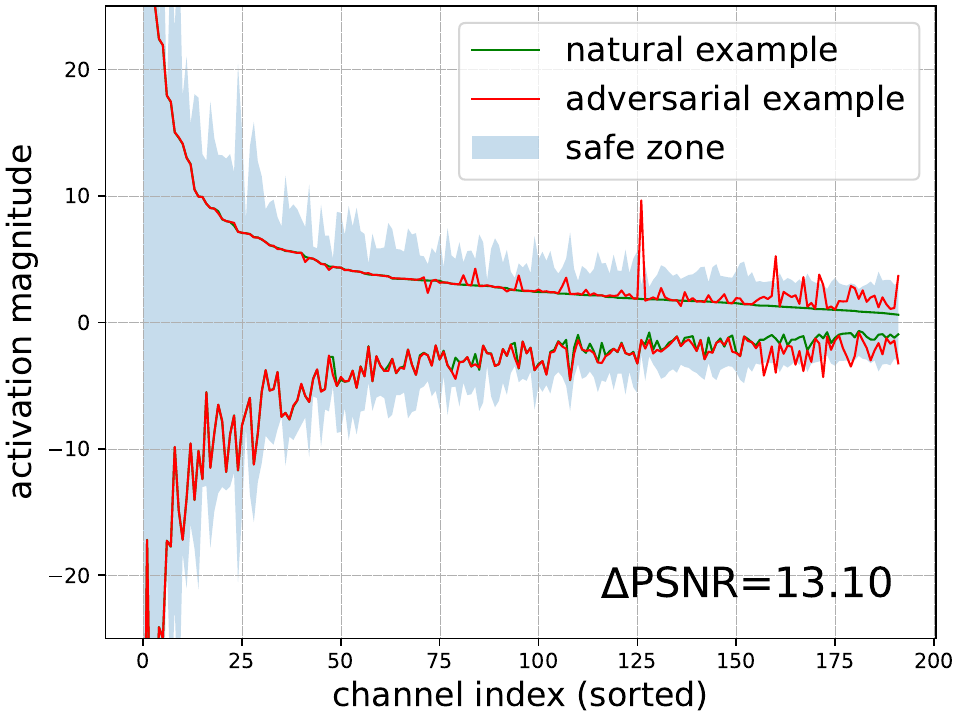}}
\subcaptionbox{AT (MSE opt.)\label{subfig:activation_d}}{\includegraphics[width=0.24\textwidth]{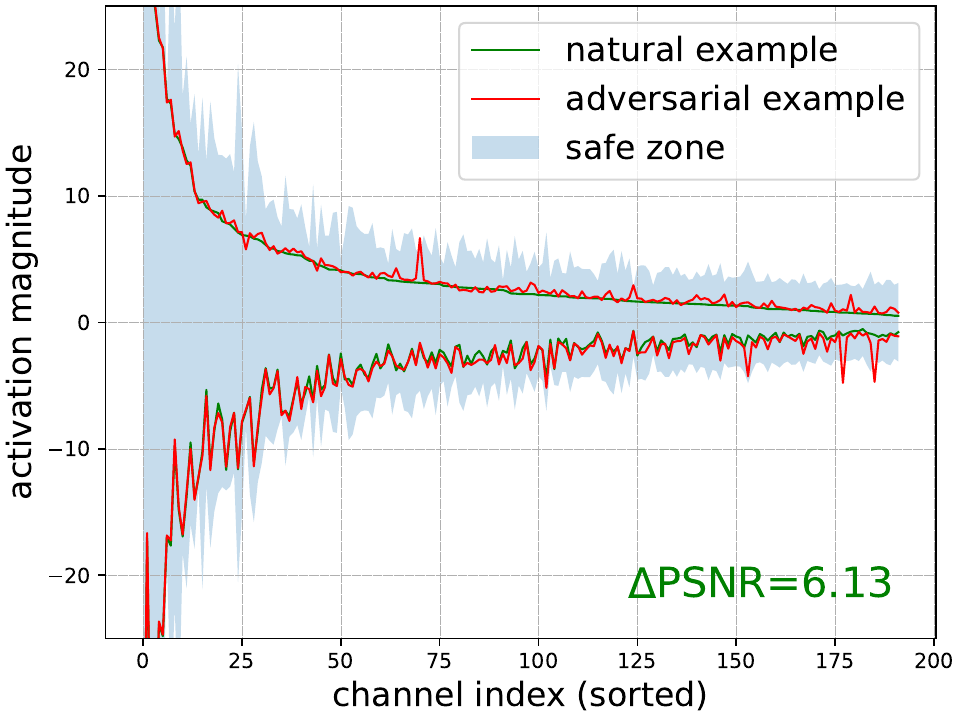}}
 \caption{\add{The magnitude range of channel-wise activation at the bottleneck layer for both the original baseline model and adversarially retrained model. Both MSE and MS-SSIM optimized Ball\'e2018 models are tested and `Kodim09' is used as an example here as the natural image and also for generating the corresponding adversarial example. The channels are sorted in descending order of the activation magnitude of the natural example.}} 
\label{fig:activation}
\end{figure*}

 \add{As also listed in Table~\ref{tab:nics}, besides the degradation in reconstruction quality, the compressed bitrate is generally increased for adversarial examples, which reveals that the injected noise is more difficult to model for efficient compression. 
 \begin{align}
    R = -\sum_{\mathbf{\hat{z}}} p(\mathbf{\hat{z}})\log_2 q(\mathbf{\hat{z}})
\label{eq:ce}
\end{align}
 We thus visualized the actual distribution of the latent features $p(\mathbf{\hat{z}})$ and the corresponding distribution $q(\mathbf{\hat{z}})$ predicted by the entropy estimator in Fig.~\ref{fig:entropy_baseline}. The final bitrate $R$ is determined by the cross entropy between $p(\mathbf{\hat{z}})$ and $q(\mathbf{\hat{z}})$ as in~\eqref{eq:ce}. As seen, the actual distribution of the adversarial example diverges significantly from the predicted distribution, yielding inaccurate entropy modeling and bitrate increase.}

\subsubsection{NIC settings}
Different from existing rules-based JPEG, VVC Intra, etc, the learning-based NIC solutions usually train separate models for different quality metrics and target bitrates.  

\begin{figure}[t]
\centering
% \subcaptionbox{Visual comparison}{\includegraphics[width=0.49\textwidth]{./figs/bitrates-new.pdf}}
\includegraphics[width=0.49\textwidth]{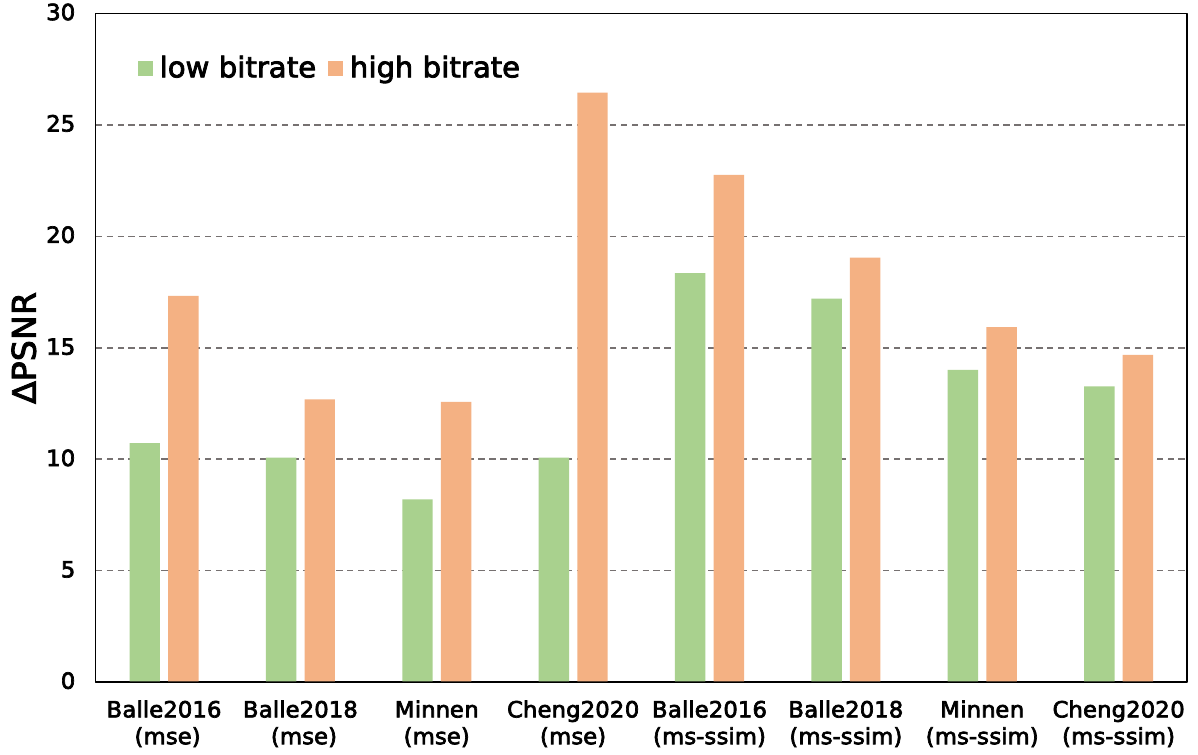}
\caption{\add{{\bf Impact of Quality Scales and Loss functions Used in NICs.} 
The "low bitrate" and "high bitrate" separately represent averaged results of models for low-quality scales ($q$ = 1,2,3) and high-quality scales ($q$  = 4,5,6). More information about the setting of the quality scale $q$ can be found in Table~\ref{tab:quality}.
}}
\label{fig:bitrate}
\end{figure}

\textbf{Loss functions.} The MSE and MS-SSIM are two commonly used quality metrics in image compression. They have also been widely applied as loss functions for training NIC models.
% \edit{One advantage of the learning-based NIC solution to existing rules-based JPEG, VVC Intra, etc, is that we can easily adapt specific loss functions towards designated quality optimization. In addition to the traditional MSE (Mean squared error) and MS-SSIM (Structural similarity) that are extensively used in model training, recent works have suggested that LPIPS, feature loss, as well as the GAN loss can be applied separately or jointly for better perceptual quality of decoded reconstructions because these metrics show better correlation with the subjective sensation~\cite{zhang2018unreasonable,huang2019extreme}.}
% We separately optimize the Ball\'e 2018~\cite{balle2018variational} using three popular loss functions, e.g., MSE, MS-SSIM, and LPIPS. As illustrated in Fig.~\ref{fig:lambda_comparison}, model instability incurred distortions are again presented, besides the impairments observed in Fig.~\ref{fig:anchors} for image coded by GAN loss optimized HiFiC,
As in Table~\ref{tab:nics}, we test various MSE and MS-SSIM optimized methods and mixed loss function (MSE+LPIPS+GAN) based HiFiC, exhibiting successful attacks regardless of the loss function used in model optimization. 

\add{When comparing MSE and MS-SSIM optimized models from the perspective of latent activations in Fig.~\ref{subfig:activation_a} and Fig.~\ref{subfig:activation_c}, we can clearly see that the activations of compressed adversarial example using the MS-SSIM optimized model vary more significantly  than that using the MSE optimized model, which is consistent with the overall measurement of $\Delta$PSNR in Table~\ref{tab:nics} that MS-SSIM optimized models are more vulnerable to adversarial attacks in most cases.}

% \begin{figure}[t]
% \centering
% \subcaptionbox{MS-SSIM opt.\label{subfig:activation_baseline_a}}{\includegraphics[width=0.24\textwidth]{./figs/activations_pretrained_hyper_sim5.pdf}}
% \subcaptionbox{MSE opt.\label{subfig:activation_baseline_b}}{\includegraphics[width=0.24\textwidth]{./figs/activations_pretrained_hyper_mse5.pdf}}
%  \caption{{\bf Channel-wise Activation Magnitude Comparison.} Both MSE and MS-SSIM optimized Ball\'e2018 models are tested and `Kodim09' is used as an example here as the natural image and also for generating the corresponding adversarial example.}
% \label{fig:activation_baseline}
% \end{figure}

% Sharp drops of PSNR and MS-SSIM are noted when quantitatively comparing the reconstruction of compressed adversarial examples to those of compressed inputs without adversarial attack. 

\textbf{Quality Scales.} In practice, images will be encoded at various quality scales (or bit rates) to satisfy diverse application requests.
Generally, a larger bit rate would come with better reconstruction quality.
%As a rate-distortion trade-off procedure, the reconstruction distortion is not the only metric when measuring the overall performance of a image compression method. Another essential aspect is the bitrate (the storage or bandwidth required to store and transmit the compressed image bitstream). Theoretically, with the bitrate getting bigger, more information of the original input image can be preserved and reconstructed at the decoder side. For a conventional codec like JPEG, it usually means preciser quantization step so the quantization error introduced would be smaller. Similarly, in the case of neural image codec, a reasonable expectation is that the adversarial attacks should be more difficult for the compression networks trained for high bitrates since they are able to produce reconstruction with much less noise. However, 
However, as shown in Fig.~\ref{fig:bitrate}, the NIC models are constantly vulnerable to adversarial attacks at all quality scales. 
% \edit{This agrees with the observations in~\cite{helminger2021lossy} that distortion exist for adversarial samples encoded at arbitrary bitrates.} 
Actually, due to the irreversibility of VAE-based methods as mentioned in~\eqref{eq:irreversible_neural_tramsform}, there exists a distortion up-bound or \textbf{AE limit} $\varphi$ in~\eqref{eq:aelimit} as mentioned in~\cite{helminger2021lossy} no matter how the bitrate $r$ increases, i.e.,
 \begin{equation}
    {\lim_{r \to +\infty}}||x - f_D(f_E(x))|| = \cancelto{\varphi}{0}.
\label{eq:aelimit}    
\end{equation}
It explains why it is applicable to generate adversarial examples for models trained for arbitrary bitrates.

Along with the increase of compressed bitrate, the quality of adversarial reconstruction gets even worse. We believe that it is because, at low bitrates, even though adversarial examples are with the perturbation, a higher quantization level would remove more details as well as noises,  which could partially stop the noise accumulation and present fewer distortions. However, when the bitrate gets higher, the model tries to retain more information of the input image, which may also accumulate the perturbation layer by layer to produce severer distortion.

\begin{figure}[h]
  % Left figure
    \centering
    \includegraphics[width=0.5\textwidth]{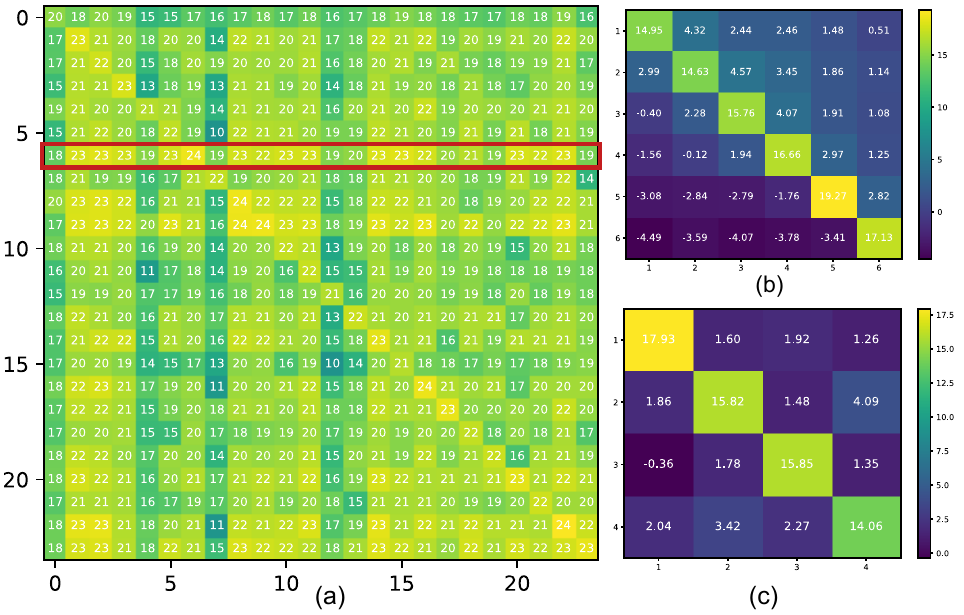}
    \caption{{\bf Attack transferability measured by $\Delta$PNSR.} \textbf{(a) Transferability across images.} Taking ``Kodim07.png'' as an example (highlighted in a red rectangle), adversarial perturbations generated on Kodim07 (the sixth row) can also cause a similar level of PSNR degradation when being directly augmented on other images. Both indices along with the $x$- or $y$-axis indicate the image ID in Kodak.  \textbf{(b) Transferability across qualities.} Models with the same Ball\'e2018 network architecture but trained for different quality scales are tested. Indices indicate the quality scales (q=1-6).  \textbf{(c) Transferability across different architectures} (0-Ball\'e2016, 1-Ball\'e2018, 2-Minnen, 3-Cheng2020). For (b) and (c), $\Delta$PSNRs over Kodak images are presented.}
    \label{fig:transfer}
\end{figure}

\textbf{Attack Transferability.}
Another area that has received extensive attention is the transferability~\cite{li2019cross, li2020towards, li2022learning, yang2020adversarial} of adversarial perturbations. In Fig.~\ref{fig:transfer} we present some preliminary tests of the transferability. For cross-image transferability, specifically, the adversarial perturbations generated for a specific image are directly added to other images to test the transferability of the adversarial perturbation. Noticeable distortions are consistently presented in adversarial reconstructions with $\Delta$PSNR measurement shown in Fig.~\ref{fig:transfer}(a), which means it is possible to generate universal perturbations that can widely affect the rate-distortion performance of multiple samples. 
Besides the transferability of adversarial perturbation across different images, there are many other aspects worth deep investigation, such as the transferability across different networks, which we also present in Fig.~\ref{fig:transfer}b-c. The results indicate that the transferability across different models is relatively small. This is mainly because our approach was not specifically designed for cross-model transferability. Nevertheless, we believe that exploring strategies to improve cross-model transferability is an interesting and challenging direction for future research. 

% \begin{figure}[t]
%     \centering
%     \subcaptionbox{cross images}{\includegraphics[width=3cm]{figs/hyper_8_ms-ssim_transfer.pdf}}%
%     \subcaptionbox{cross model}{\includegraphics[width=3cm]{figs/transferability_matrix.pdf}}%
%     \subcaptionbox{cross model}{\includegraphics[width=3cm]{figs/transferability_matrix_cross_method.pdf}}
%     \caption{{\bf Attack transferability measured by $\Delta$PNSR upon Kodak dataset.} Taking ``Kodim07.png'' as an example (highlighted in a red rectangle), adversarial perturbations generated on Kodim07 (the sixth row) can also cause a similar level of PSNR degradation when being directly augmented on other images. Both indices along with the $x$- or $y$-axis indicate the image ID in Kodak.}
%     \label{fig:transfer}
% \end{figure}

\section{Attack Defense}
\label{sec:attack_defense}
Extensive experiments in the last section report that existing NIC models are vulnerable to adversarial attacks, despite different network structures and settings.
In this section, we will explore possible defense strategies against adversarial attacks. 

\subsection{Defense Strategies} \label{sec:data_augmentation}
\textbf{Pre-processing.} Some  pre-processing based defense strategies such as random resizing and JPEG compression would degrade the coding performance even without attack, making them not suitable for image compression tasks (see Fig.~\ref{fig:bitdepth} and Fig.~\ref{fig:rd}). However, there still exist certain transformations such as flipping and rotations at specific angles ($90^\circ$, $180^\circ$, $270^\circ$) that do not have negligible impact on the coding performance without attack. 
Also, such geometric transformations have been proven to be able to mitigate the effect of adversarial perturbation~\cite{thang_image_2019, tian2018detecting, choi_deep_2021}. Therefore we choose the geometric self-ensemble~\cite{lim_enhanced_2017} as one of the defense strategies for evaluation. 

Geometric self-ensemble (ensemble for short) was first proposed to improve the performance of super-resolution networks.
It generates several augmented adversarial inputs $x_i^* = T_i(x)$ for each input image by flipping and rotation, where $T_i$ represents eight geometric transformations in total. With these augmented inputs, we can generate corresponding reconstruction images $\{\hat{x}_{1}^*,\ldots, \hat{x}_{8}^*\}$ using the NIC networks. We then apply inverse transform to those output images to get  $\tilde{x}_i^* = T^{-1}_{i} (\hat{x}_{i}^*)$. Finally, we select the transformed output with minimum distortion compared with the original input $x$ as the final self-ensemble result. Besides, an extra transformation index $i$ needs to be stored in the bitstream to help the decoder to apply the correct inverse transform.

\textbf{Adversarial Training.}
\add{Note that preprocessing (transformation) based defense strategies only apply process on the input images while keeping the models unchanged. 
% \edit{They can usually be countered by designing stronger attacks.}
% (see Fig.~\ref{fig:ensemble} in Appendix).} 
Besides, multiple transformed input images need to be compressed and reconstructed to find the optimal output, making the computational 
complexity multiple times greater. 
Therefore, the robustness improvement of the model itself is also highly worth investigating.} 

Existing NIC approaches often use noise-free (or at least less noise), high-quality, uncompressed image datasets to train respective models. They work well on clean and uncompressed content but are possibly vulnerable to noisy inputs as aforementioned.
%\edit{Recalling that noisy inputs can be simulated by adversarial example generation, we then propose to derive adversarial examples using original, random samples in original training dataset, by which we can basically enrich the data distribution for improving the model generalization afterwards.} 
Experiments in previous sections have shown that such noisy inputs can be simulated by adversarial example generation, we then propose to utilize these generated adversarial examples for model finetuning, by which we can enrich the distribution of training data and effectively improve the model generalization.

\begin{algorithm}[t] 
\caption{Adversarial Training}
\hspace*{0.02in} {\bf Input:} 
\hspace*{0.02in}
Encoder $f_E()$ parameters $\theta$, Decoder $f_D()$ parameters $\phi$, Dataset $\mathbb{D}$, finetuning iterations $N$, adversarial example generation iterations $M$, RDO penalty $\lambda$;\\
\hspace*{0.02in} {\bf Output:}
$\theta^*$, $\phi^*$
\begin{algorithmic}[1]
\STATE $\theta^*, \phi^* =\theta, \phi$ //initialize using pretrained baseline model
% \STATE $i$ = 0
\FOR{i $<$ $N$}
    %\STATE randomly sample $x_j$ from $\mathbb{D}$
    \STATE $\Vec{\bf x}_{\bf ori}$ = RANDOM\_SAMPLE\_FROM($\mathbb{D}$)
    \STATE {$\bf n$ = RANDOM\_NOISE(), $\Vec{\bf x}^*$ = $\Vec{\bf x}_{\bf ori}$ + n}
    \FOR{$j \leq$ $M$}
        \STATE optimize $\Vec{\bf x}^*$ by Eq.~\eqref{eq:dloss}
        \STATE $j$ = $j$ + 1
    \ENDFOR
    % \STATE $\Vec{\bf x}_{\bf new}$ = [$\Vec{\bf x}_{\bf ori}$, $\Vec{\bf x}^*$]
    %\STATE $\mathbb{D}_{adv}$ = [${x_0^*, x_1^*, x_2^*, ..., x_M^*}$]
    %\STATE $\mathbb{D}_{new}$ = [$\mathbb{D}_{ori}, \mathbb{D}_{adv}$]
    %\WHILE{training}
% 　　\IF{condition}
    %\STATE loss = Eq.~\eqref{eq:vae_rd}
    %\STATE $\Theta^* =  \Theta^* - \delta\frac{d loss}{d \Theta}$
    \STATE $dloss$ = DISTORTION\_FUNCTION($f_{D}(f_{E}(\Vec{\bf x}^*))$, $\Vec{\bf x}^*$)
    \STATE $rloss$ = RATE\_FUNCTION($f_{E}$($\Vec{\bf x}^*$))
    \STATE $loss$ = $\lambda dloss + rloss$
% 　　\STATE update $\Theta^*$ and $\Phi^*$ %$\mathbb{D}_{new}$
    \STATE $\theta^*$ = OPTIMIZER($\theta^*, loss$)
    \STATE $\phi^*$ = OPTIMIZER($\phi^*, loss$)
    %$\mathbb{D}_{new}$
    %\ENDWHILE
    \STATE $i$ = $i$ + 1
\ENDFOR
\STATE \RETURN $\theta^*$, $\phi^*$
\end{algorithmic}
\label{method:iter}
\end{algorithm}

As described in Algorithm~\ref{method:iter}, we iteratively implement the attack-and-finetune strategy to update the pretrained model.
Here $\theta$, $\phi$ are trainable parameters in encoder $f_E()$ and decoder $f_D()$ respectively. 
$N$ is the total iterations of adversarial training and in each iteration, a batch of 8 images are randomly sampled from the training dataset ${\mathbb{D}}$. 
% \edit{in which four of them are kept as original images $\Vec{\bf x}_{\bf ori}$ and the other four images would be used to generate adversarial examples $\Vec{\bf x}^*$ with $M$ steps of optimization following~\eqref{eq:dloss}.}
These original images $\Vec{\bf x}_{\bf ori}$ will be used to generate adversarial examples $\Vec{\bf x}^*$ with $M$ steps of optimization following~\eqref{eq:dloss}, and then $\Vec{\bf x}^*$ will be fed into the network to update the pretrained models with Adam optimizer. In this paper, we set $N$ = 1000, $M$ = 300. 
Randomly-cropped 256$\times$256$\times$3 patches from Vimeo90K\cite{xue2019video} are used as the training baseline dataset ${\mathbb{D}}$. 
% \delete{We then randomly subsample from this original dataset to generate the adversarial examples following the \eqref{eq:dloss}.}
Note that $l_2$ distance is used and two noise threshold level settings ($\epsilon = 1\text{e}^{-3}$ and $1\text{e}^{-4}$) are applied and compared. The generation of each batch of adversarial examples costs about 22s on the platform with Intel Xeon Silver 4210 CPU@2.20GHz and a single NVIDIA RTX 2080Ti GPU.

\subsection{Defense Efficiency}

%\edit{The pretrained Ball\'e 2018~\cite{balle2018variational} models are refined on this assembled new dataset ${\sf D}_{new} = \{{\sf D}, {\sf D}_{adv}\}$.}

We mainly show the efficiency of attack defense in two folds: one aspect is about the robustness improvement both visually and quantitatively; the other one is about the coding efficiency of the NIC models with the defense algorithm integrated. \add{Here, the Ball\'e2018 method is first used as an example to evaluate the impact of different defense strategies on defense efficiency in Fig.~\ref{fig:at_vi}-\ref{fig:rd}. More comprehensive results for different models and settings can be found in Table~\ref{tab:nics_defense}}. 

\add{
\begin{figure}[t]
    \centering 
    \includegraphics[width=0.49\textwidth]{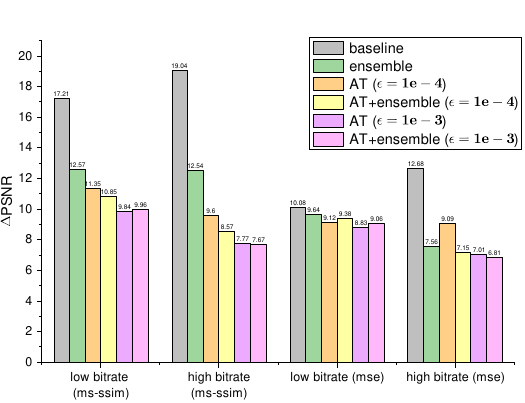} 
    \caption{\add{{\bf Robustness Improvement Comparison.} Baseline models (baseline) are pretrained NIC models, provided by publicly available resources~\cite{begaint2020compressai}. The ensemble strategy is also directly applied to the pretrained baseline models. Adversarial retrained models (AT) are finetuned from the baseline models with two perturbation levels ($\epsilon$=1$\text{e}^{-4}$ and $\epsilon$=1$\text{e}^{-3} $). The AT+ensemble means applying the self-ensemble strategy to the AT models. The results are averaged across all Kodak images. The ``low bitrate" and ``high bitrate" separately represent averaged results of models for low-quality scales ($q$ = 1, 2, 3) and high-quality scales ($q$ = 4, 5, 6) in Table~\ref{tab:quality}.}}
    \label{fig:at_vi}
\end{figure}
}

\subsubsection{Robustness Improvement}
\begin{figure}[t]
\centering
\includegraphics[width=0.49\textwidth]{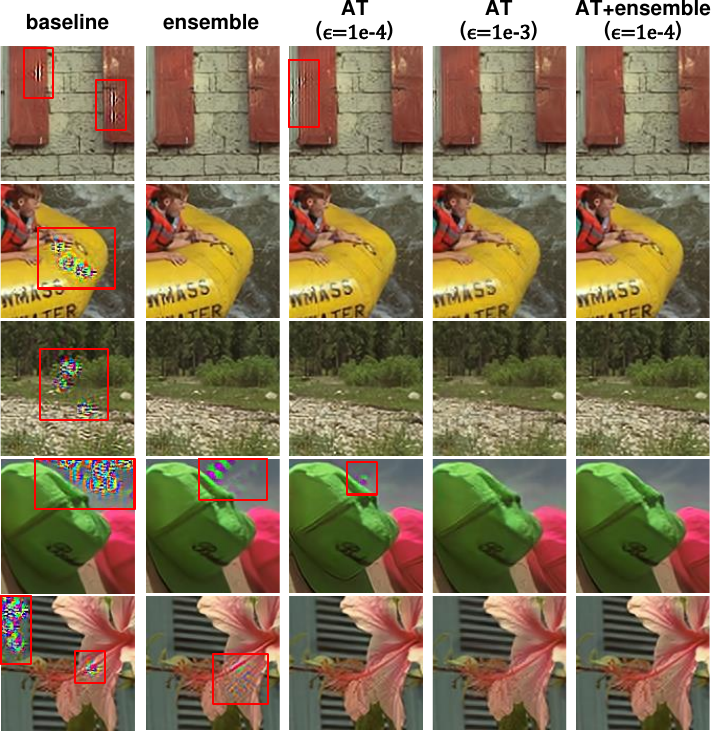}
\caption{\add{Visual comparison of the defense efficiency of different strategies. The tested models here are MS-SSIM optimized Ball\'e2018 with quality scale $q$ = 5.}}
\label{fig:at}
\end{figure}

\add{In Fig.~\ref{fig:at_vi}, the defense efficiency measured by $\Delta \text{PSNR}$ is displayed. 
The \textbf{ensemble} strategy, two input perturbation threshold levels during adversarial training (\textbf{AT}), and their combination (\textbf{AT+ensemble}) are tested and compared. Note that the AT ($\epsilon$ = 1$\text{e}^{-4}$ and $\epsilon$ = 1$\text{e}^{-3}$) here only indicates  the perturbation levels used in training. While testing the defense efficiency, the input perturbation is consistently set to 1e-4.

 Experiments clearly show that both the self-ensemble strategy and adversarial training can effectively improve the robustness of underlying models and usually AT models can achieve better defense performance compared with the self-ensemble strategy.
 For AT models, more robustness improvement can be achieved with a larger perturbation level $\epsilon$ during adversarial training. Besides, when $\epsilon$ during training is set low ($\epsilon$ = 1e-4), the combination of AT and self ensemble can further improve defense efficiency, while when $\epsilon$ is already high, the additional application of self ensemble can no longer provide an obvious performance gain. 
 Figure~\ref{fig:at} further visualizes the reconstructions of adversarial examples with different defense strategies. It is also well observed that the distortions caused by the adversarial attack are effectively suppressed.}
 
 % Compared with the model before retraining, adversarially-finetuned NIC model can effectively defend against perturbation augmented in the input image, which we believe is because the pretrained NIC model is improved by adversarial training to characterize the distribution of noisy input better. 
 % We have also featured some imperfect examples in Fig.~\ref{fig:at}, where AT models do not completely eliminate but still significantly suppress the distortion.

\add{In addition to the above subjective and objective comparisons, Fig.~\ref{fig:activation} shows the activation magnitude comparison between natural images and adversarial examples for both the baseline model and adversarially retrained model. For model trained on natural images in Fig.~\ref{subfig:activation_a} and~\ref{subfig:activation_c}, the activation of adversarial examples shows distinct changes compared with natural images. While for AT models retrained on adversarial examples, the activation inconsistency in Fig.~\ref{subfig:activation_b} and~\ref{subfig:activation_d} is greatly alleviated. Previous work on the defense of image classification networks~\cite{bai_improving_2022} has also presented similar phenomena. Besides, compared with the distribution of the baseline model shown in Fig.~\ref{fig:entropy_baseline}, Here in Fig.~\ref{fig:entropy_at} for the adversarially trained model, the divergence between the predicted distribution and actual distribution is greatly alleviated. As a result, the bitrate gain caused by adversarial perturbation can also be better controlled now. More experimental results on different methods can be found in Table~\ref{tab:nics_defense}.

% Here we choose the channel that is spatially most correlated with the output distortion. \edit{Noticeable variations in features confirm the insufficient capacity of native, pretrained baseline model to model and reconstruct these pertubations.} 
% Specifically, Fig.~\ref{fig:activation} gives the overall distribution comparison in latent feature space between original input and adversarial input using the baseline model. As we can see, feature distributions are almost the same for some channels like the 17-th, 37-th and 149-th channels, while they are very different for some channels like the 41-st, 124-th and 136-th channels, demonstrating the capability of the adversarial examples to enrich the distribution of the original dataset.
% Therefore, by finetuning the baseline model with these augmented adversarial examples, 
% we can extend the model representative capacity to effectively reconstruct the perturbation.} 
%\edit{As shown in Fig.~\ref{fig:augment_c}, \delete{with similar response in latent features as in Fig.~\ref{fig:augment_b}, the reconstruction distortion is greatly alleviated.}} 
%\delete{As a result, the feature distribution is refined correspondingly as in Fig.~\ref{fig:distribution}.}

% \delete{Besides, implying that the characterization of noise perturbation in input images can be well represented using responses of a set of feature channels. Precisely connecting noise perturbation with specific channels in latent space is an interesting topic for our future study.}
}

\begin{table}[htbp]
\centering
\caption{\add{The $\Delta$bpp and $\Delta$PSNR of adversarial examples compared with original images for different defense strategies. For a fair comparison, the AT models here are AT($\epsilon$=1e-4) since this setting would not degrade the original coding efficiency. Results are averaged over all 6 quality levels and both mse and ms-ssim optimized models on the Kodak dataset. Lower $\Delta$bpp and $\Delta$PSNR mean better defense efficiency.}}
\label{tab:nics_defense}
\begin{tabular}{c|l|c|c}
\hline
\textbf{Methods} & \textbf{Defense Strategy}  & \textbf{$\Delta$bpp}$\downarrow$ & \textbf{$\Delta$PSNR}$\downarrow$                        \\
\hline
                            & baseline (no defense) & 4.74\%   & 17.29                       \\
                            & ensemble         & 3.66\%   & 12.73                       \\
                            
                            & AT                    & \textbf{2.23\%}   & 11.54                       \\
\multirow{-4}{*}{Balle2016} & \textbf{AT+ensemble}      & 3.15\%   & \textbf{8.98}                        \\ \hline
                            & baseline (no defense)              & 8.37\%  & 13.80                      \\
                            & ensemble         & 7.62\%   & 8.60                         \\
                            & AT                    & 6.51\%   & 9.79                       \\                
\multirow{-4}{*}{Balle2018} & \textbf{AT+ensemble}      & \textbf{5.63\%}   & \textbf{8.27}                        \\ \hline
                            & baseline (no defense)             & 10.46\%  & 12.68                      \\
                            & ensemble         & \textbf{9.27\%}   & 10.86                       \\
                            & AT                    & 9.89\%   & 8.90                       \\
\multirow{-4}{*}{Minnen}    & \textbf{AT+ensemble}      & 9.32\%   & \textbf{8.62}                        \\ \hline
                            & baseline (no defense)              & 62.90\%  & 16.13\\
                            & ensemble         & 9.26\%   & 10.82                       \\
                            & AT                    & 12.60\%  & 10.60                        \\
\multirow{-4}{*}{Cheng2020}     & \textbf{AT+ensemble}      & \textbf{9.01\%}   & \textbf{9.14}\\
\hline
\end{tabular}
\end{table}

\begin{figure}
    \centering
    \includegraphics[width=8.8cm]{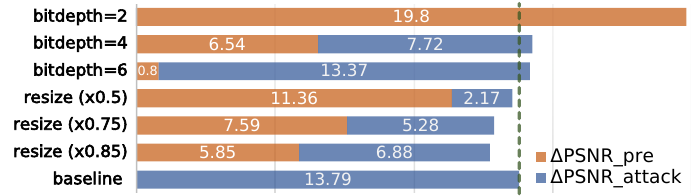}
    \caption{{\bf Defense efficiency using different pre-processing strategies}. $\Delta \text{PSNR}_{pre}$ indicates the PSNR degradation directly caused by the pre-processing strategies like resizing or bit depth reduction. $\Delta \text{PSNR}_{attack}$ is the further quality degradation caused by adversarial attack after applying the defense. The overall quality degradation equals the sum of $\Delta \text{PSNR}_{pre}$ and $\Delta \text{PSNR}_{attack}$. Here pretrained Ball\'e2018 models are tested and used as the baseline.}
    \label{fig:bitdepth}
\end{figure}

Another two preprocessing-based defense strategies (spatial resizing and bit depth reduction) are also included for comparison. For the methods using bit-depth reduction, the original input image with a bit depth equal to 8 is truncated to lower bit depths, e.g., 2, 4, 6,  to reduce the impact of adversarial perturbations. Similarly, spatial resizing  downsamples the original image to a smaller size at a factor of 0.5, 0.75, or 0.85 and correspondingly upsamples them to the original resolution for future applications. As shown in Fig.~\ref{fig:bitdepth}, these methods can alleviate the PSNR degradation incurred by the adversarial perturbation. However, both spatial resizing and bit depth reduction in pre-processing inevitably introduce additional quality degradation, leading to similar or even worse overall degradation when compared with the anchor using baseline models without applying any defense strategies at all.

% \begin{table}[]
% \begin{tabular}{lcccc}
% \hline
%                                & $\Delta$\textbf{bpp} & $\Delta\textbf{PSNR}_{all}$ & $\Delta\textbf{PSNR}_{pre}$ & $\Delta\textbf{PSNR}^'$ \\\hline
% \textbf{baseline (bitdepth=8)} & 8.37\%        & 13.79            & 0                 & 13.79          \\\hline
% \textbf{bitdepth=6}            & 12.36\%       & 14.17            & 0.8               & 13.37          \\\hline
% \textbf{bitdepth=4}            & 19.28\%       & 14.26            & 6.54              & 7.72           \\\hline
% \textbf{bitdepth=2}            & 54.03\%              &                 18.78
%  & 19.8
%                   & -1.02\\\hline
% \end{tabular}
% \caption{Performance of bit depth reduction defense strategy. $\Delta\textbf{PSNR}^'$ equals $\Delta\textbf{PSNR}_{all} - \Delta\textbf{PSNR}_{pre}$.}
% \label{tab:pre_defense}
% \end{table}

%failed to repeat the generation of new adversarial examples on these retrained models, which means that they are now also robust to new distortion attacks. With these adversarial examples of previously unseen distribution, the compression network now can be refined to be much robuster. }

%As shown in Fig.~\ref{fig:distribution}, the latent feature distribution in different channels shows obvious correlation and difference between these two datasets, demonstrating the capability of the adversarial examples to enrich the original dataset.
%Fig.~\ref{fig:augment} further shows the visual comparison of the adversarial examples, latent features $\bf z$ and reconstructions before and after adversarial retraining. 
%In Fig.~\ref{fig:augment_b}, with the generated noise added in the input image, 
%the latent feature representation  in this channel is obviously different from the latent features of the original image in Fig.~\ref{fig:augment_a}, which also matches with the result in Fig.~\ref{fig:distribution}. 
%Meanwhile, obvious distortion is also observed in the reconstruction at same spatial location as the latent features, showing that the changes in these channels are the main culprit causing the final reconstruction distortion. 
%With these adversarial examples of previously unseen distribution, the compression network now can be refined to be much robuster. 
\add{
\begin{figure}[t]
    \centering 
    \subcaptionbox{MSE opt.}{\includegraphics[scale=0.245]{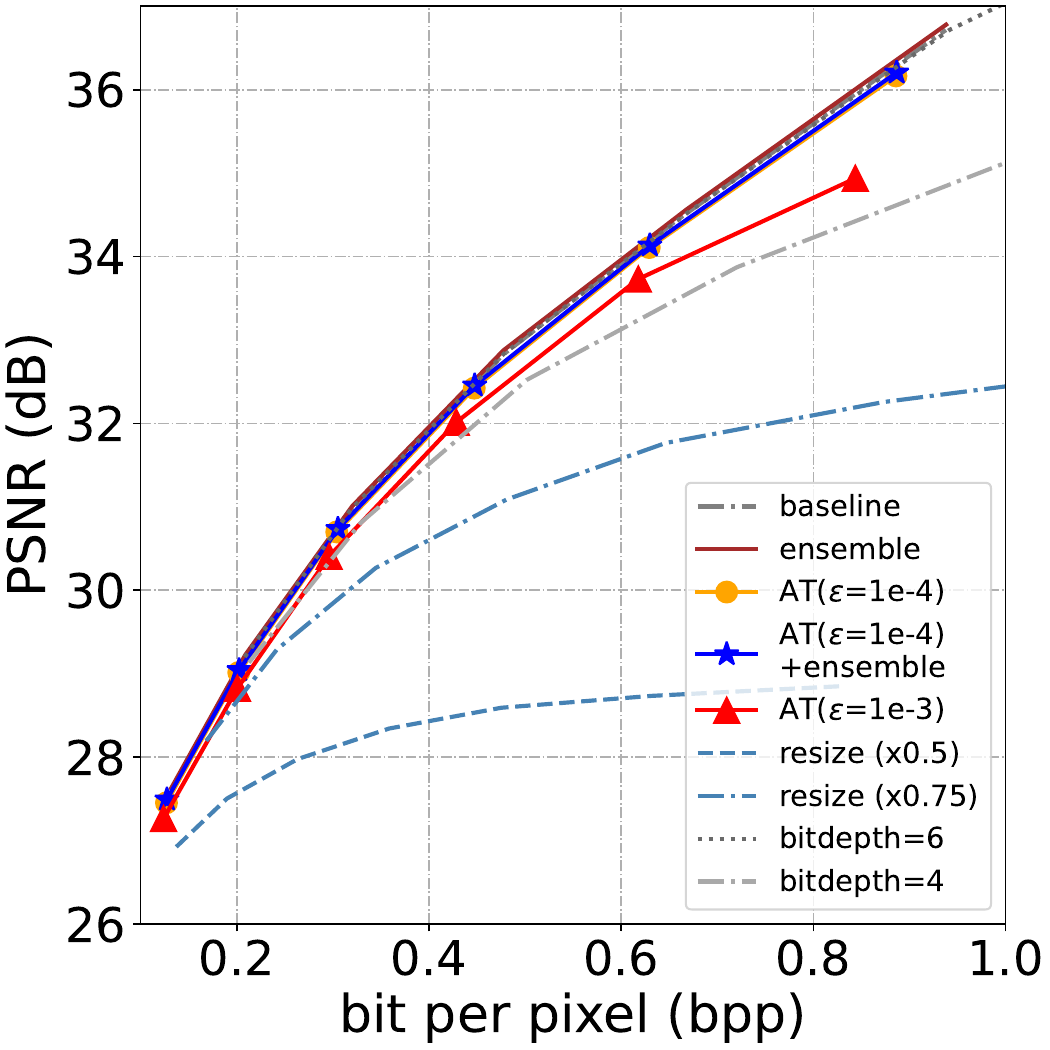}} 
    \subcaptionbox{MS-SSIM opt.}{\includegraphics[scale=0.245]{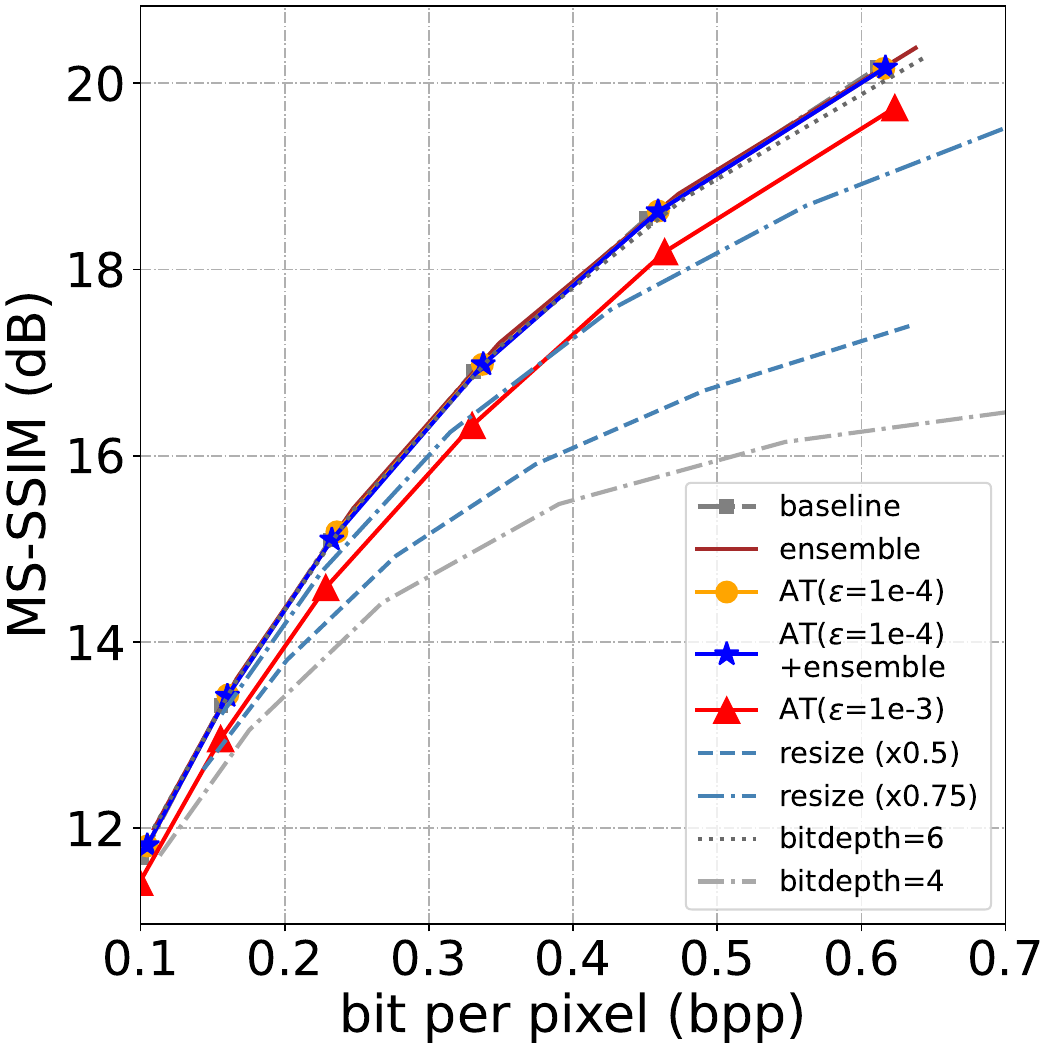}}
     \caption{{\bf Coding Efficiency.} Rate-distortion efficiency averaged over Kodak images. Ball\'e2018~\cite{balle2018variational} models are exemplified.}
    \label{fig:rd}
\end{figure}
}

\subsubsection{Coding Efficiency}
Previous discussions show that NIC models can effectively defend against attacks through defense strategies. Another important question then arises: whether the coding efficiency for original natural images is retained after applying defense strategies. Figure~\ref{fig:rd} shows the rate-distortion curves averaged over Kodak images. As aforementioned, pre-processing strategies like spatial resizing and bit depth reduction significantly degraded the performance even for clean natural images, while self-ensemble based defense strategy will not degrade the original coding performance for both baseline models as well as AT models.

For the adversarially trained models with two perturbation levels during training, %\delete{retrained baseline improves the native baseline model in coding efficiency which is mainly because of the use of different training dataset. High-quality, and high-definition image samples in DIV2K could potentially lead to a better trained compression network for inference.} 
when the perturbation level is lower ($\epsilon$=1e-4), the coding efficiency of AT models roughly aligns with the pretrained baseline models. However, as the input perturbation increases to $\epsilon$=1e-3, although the robustness can be further improved as in Fig.~\ref{fig:at}, the coding efficiency of AT models starts to deteriorate significantly, especially at high bitrates. 
Therefore, by setting appropriate input perturbation thresholds during training, the adversarial training could not only effectively improve the model robustness, but also clearly retain the coding efficiency of the original model.

In summary, these defense strategies can be easily integrated into existing learned compression codecs without requiring any changes to the underlying compression framework. Extensive experiments show that the proposed defense solutions are simple, effective, and generalizable to most existing neural image compression approaches. 

%With robustness improved, the rate-distortion performance in Fig.~\ref{fig:rd} shows no obvious drop compared with the baseline models. The results indicates that this augmentation method can be easily embedded to the training or enhancement of existing models, with almost no harm to the overall performance. Note that the performance of the original downloaded baseline model can also be improved slightly after refining on the baseline training dataset ${\sf D}_{ori}$ only (with no augmented adversarial examples) since our dataset and training setting may be different from the original model provider. For fair comparison, we provide both the performance of: \#3) the originally downloaded baseline model, \#2) refined baseline model on ${\sf D}_{ori}$ and \#1) the refined model on ${\sf D}_{new}$. 

%All experiments show that, with only a few new generated examples augmented, the robustness of the learned compression network can be significantly improved, and it is a quick plug-and-play strategy with no changes to the compression framework.
\begin{figure}[t]
    \centering 
    \includegraphics[width=0.48\textwidth]{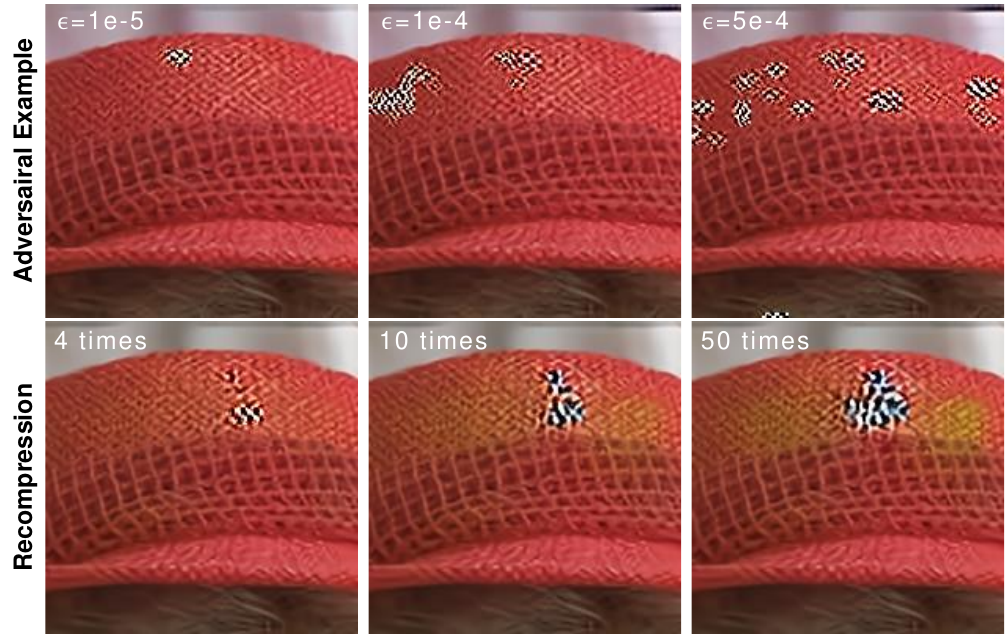} 
     \caption{\add{{\bf Visualization of similar distortion pattern caused by image recompression and adversarial perturbation.} A pretrained Ball\'e2018 baseline model is tested as an example. The first row is the reconstructions of adversarial examples generated under different perturbation levels $\epsilon$. The second row is reconstructions after different times of recompression.}}
    \label{fig:similar_distortion}
\end{figure}
\begin{figure}[t]
    \centering     \subcaptionbox{MSE opt.\label{subfig:rec_rd_psnr}}{\includegraphics[width=0.46\textwidth]{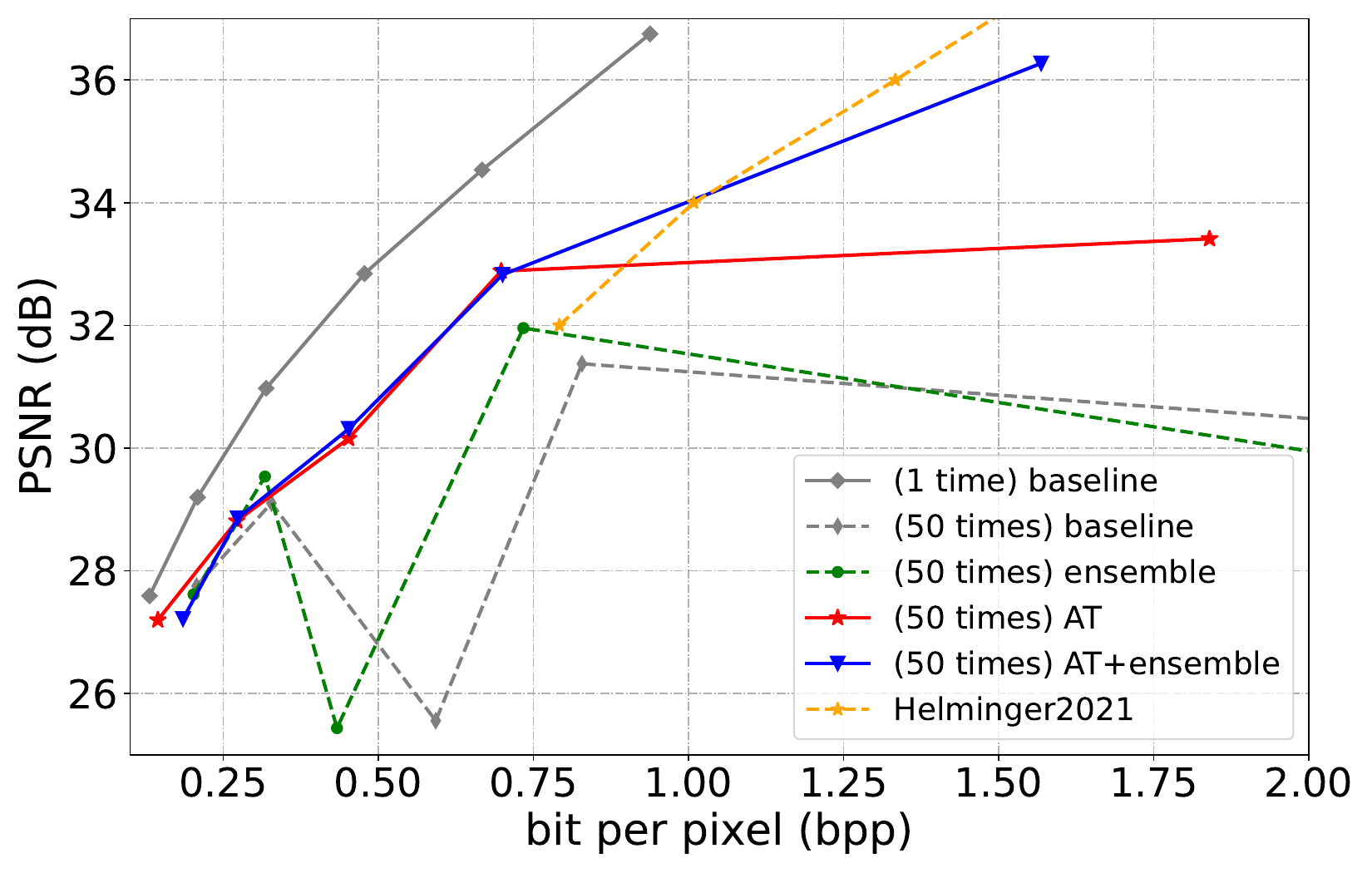}}
    \subcaptionbox{MS-SSIM opt.\label{subfig:rec_rd_msim}}{\includegraphics[width=0.46\textwidth]{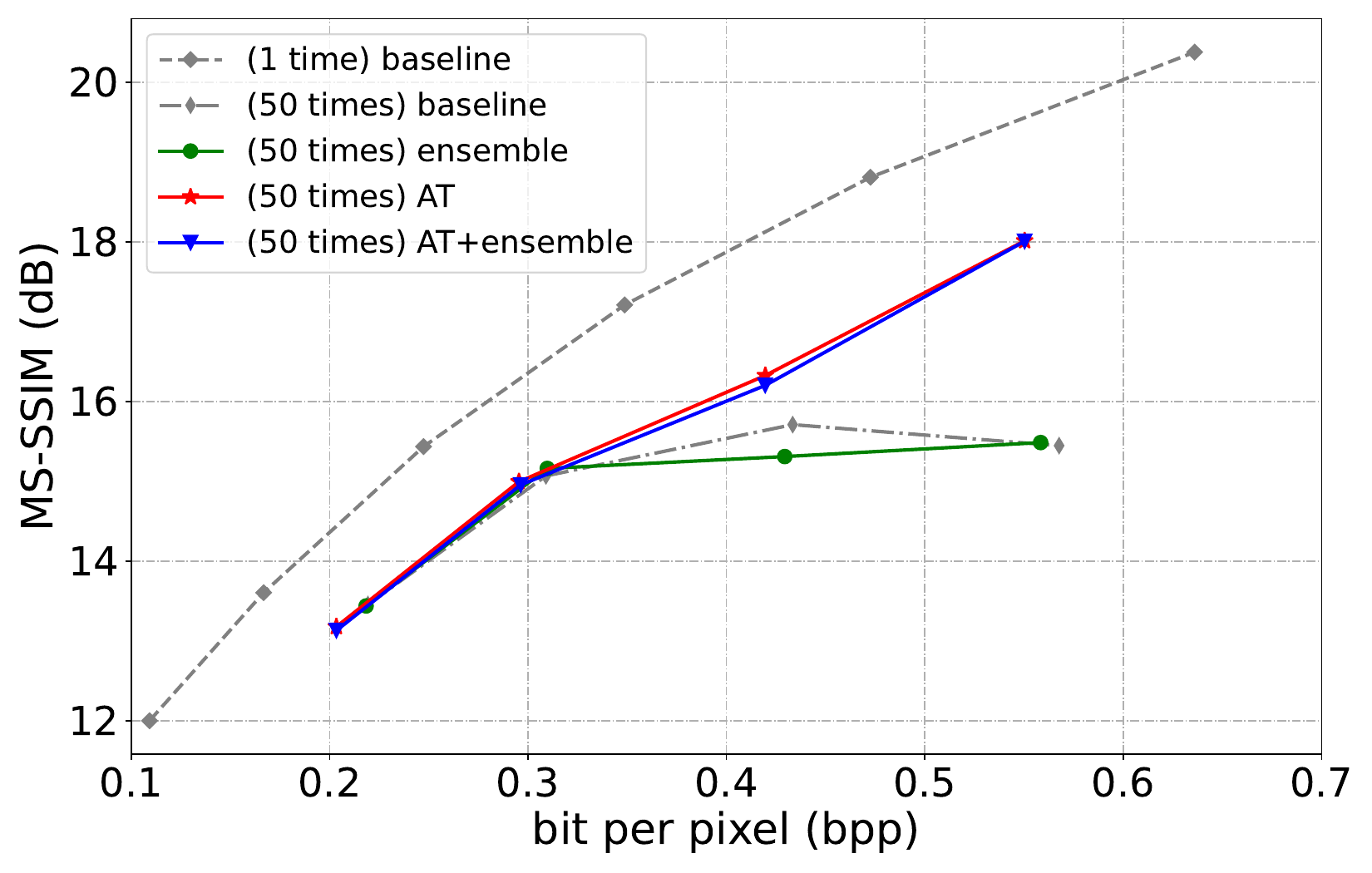}}
    % \subcaptionbox{Changes of PSNR (dB) during repetitions (Kodim04)\label{subfig:dpsnr_1}}{\includegraphics[width=0.24\textwidth]{./figs/recompression_rd.pdf}} 
    % \subcaptionbox{Kodim14\label{subfig:dpsnr_2}}{\includegraphics[width=0.24\textwidth]{./figs/recompression_rd_2.pdf}} 
     \caption{{\bf Coding efficiency degradation comparison averaged over Kodak images.} Ball\'e2018 models are tested and the AT models are trained with $\epsilon$=1e-4.}
    \label{fig:recompression_rd}
\end{figure}

\subsection{Recompression Application}

% \edit{Image recompression is very common in applications. As reported in~\cite{kim2020}, successive recompression using learnt image coder would sometimes lead to severe distortion in reconstruction.}
Looking back to the instability problem of learned image recompression mentioned at the beginning of the paper,
to tackle it, Kim {\it et. al}~\cite{kim2020} proposed to include images that were repeatedly compressed in training to improve the model robustness on purpose. Here, we want to evaluate whether our proposed defense strategies that are not designed for any specific scenario can also improve the model robustness more generally. 
% \edit{However, given that compression artifacts can also be treated as the additive noises, we then test the finetuned compression model on this recompression task.}

\add{Examples in Fig.~\ref{fig:similar_distortion} show that some of the distortions caused by recompression are of high similarity to examples generated by adversarial attacks. Such similar patterns prove that the adversarially trained models can also be generalized to practical applications like image recompression.}
The visual comparison using Kodak images is shown in Fig.~\ref{fig:recompression}. After 50 repetitions of recompression, severe distortions can be found in reconstructions encoded using the pretrained baseline model, while these unexpected distortions are greatly alleviated in decoded images that were compressed using the AT model.
% Fig.~\ref{subfig:dpsnr_1} and~\ref{subfig:dpsnr_2} show some examples of the degradation trend comparison of reconstruction quality with the growth of the number of recompression times.

\add{Figure~\ref{fig:recompression_rd} further shows the overall rate-distortion comparison for recompression at different bit rates. As seen, the performance of the baseline model degrades significantly after 50 times recompression.
We can also clearly see from the rate-distortion curves that models with higher quality scales are with severer quality degradation, which is consistent with our observation in Sec.~\ref{sec:method_distorion} by the adversarial attack. This also evidences that our proposed adversarial attack-based robustness evaluation method can really be indicative of the robustness of the NIC models in practical applications. 

For different defense strategies, the self-ensemble strategy provides limited performance improvement, while the adversarial training strategy still works well by significantly decreasing the quality degradation caused by image recompression.}
%, in which new distortion is found at a different location, showing that although the proposed method cannot always successfully deal with all possible 'bad' cases, it can still significantly improve the stability of learning-based compression methods on real scenarios. Different from the strategy proposed by Kim {\it et. al}~\cite{kim2020}.
This clearly shows the generalization of our proposed adversarial training strategy. We also expect that our methodology can be applied to other pixel domain regression tasks for robustness evaluation and improvement in future works such as quality enhancement, super-resolution, etc.
%{since our noise perturbation optimization in \eqref{eq:dloss} generally cover possible noise augmentation scenarios.}

%is not specially designed for any specific scenario but still has good performance in this experiment, showing the generalization of our proposed method not limited to image recompression but also to other potential scenarios.

\section{Conclusion}
\label{sec:conclusion}

In this paper, we propose to generate adversarial examples to examine the model robustness of existing learned image compression methods. Experiments show that all tested methods are vulnerable to adversarial attacks regardless of their network architectures, loss functions, and quality settings. To tackle it, we then explore defense strategies to improve the robustness of NIC models. Results have shown the effectiveness of our methodology, not only successfully defending the adversarial attack with great distortion alleviation in reconstruction, but also retaining the outstanding compression performance. 
It is also demonstrated that our methodology can be easily extended to the targeted attack. Our proposed method is generalizable to the most popular learned image compression frameworks and can be applied to other pixel domain regression tasks as well in future works.

\section{Acknowledgement}
We are very grateful for the authors of~\cite{balle2016end,balle2018variational,minnen2018joint,Cheng2020Learned,chen2020,mentzer2020high,xie2021enhanced} who have made their learning-based image compression methods open source to the public.

\bibliographystyle{IEEETran}
\bibliography{main}

\begin{IEEEbiography}[{\includegraphics[width=1in,height=1.35in,clip,keepaspectratio]{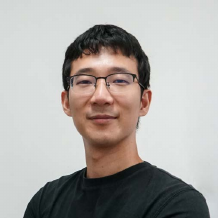}}]{Tong Chen} received his B.E and M.E. degree in Electronic Science and Engineering from Nanjing University, Jiangsu, China in 2015 and 2018. He is now pursuing the PhD. degree in Nanjing University. His current research focuses on image/video processing, including deep learning based image coding. He is a co-recipient of the  2018 PCM Best Paper Finalist, 2020 IEEE MMSP Image Compression Grand Challenge Best Performing Solution.
\end{IEEEbiography}

\begin{IEEEbiography}[{\includegraphics[width=1in,height=1.25in,clip,keepaspectratio]{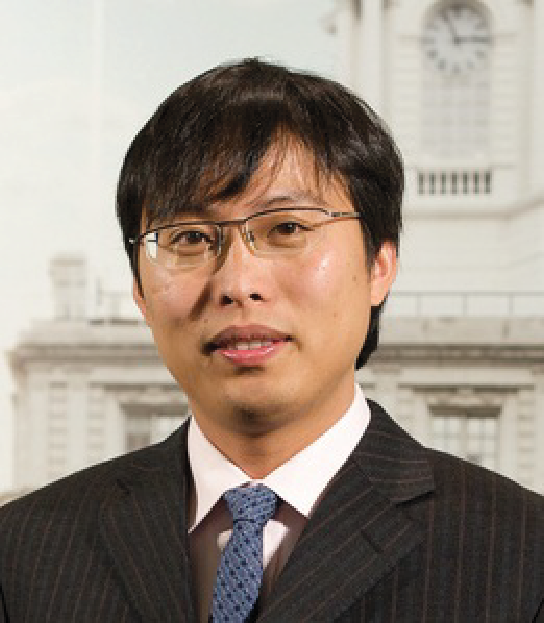}}]{Zhan~Ma} 
 is a Professor in the School of Electronic Science and Engineering, Nanjing University, Jiangsu, 210093, China. He received his Ph.D. degree from New York University, New York, in 2011, and his B.S. and M.S. from the Huazhong University of Science and Technology, Wuhan, China, in 2004 and 2006 respectively. From 2011 to 2014, he has been with Samsung Research America, Dallas, TX, and  Futurewei Technologies, Inc., Santa Clara, CA, respectively. His research focuses include learned image/video coding and computational imaging. He was awarded the 2018 PCM Best Paper Finalist, the 2019 IEEE Broadcast Technology Society Best Paper Award,  the 2020 IEEE MMSP Grand Challenge Best Image Coding Solution, and the 2023 IEEE WACV Best Algorithms Paper Award.
\end{IEEEbiography}

\clearpage
\appendix{}

\subsection{Distance Measurement}
\begin{figure}[htbp]
\centering
\includegraphics[scale=0.14]{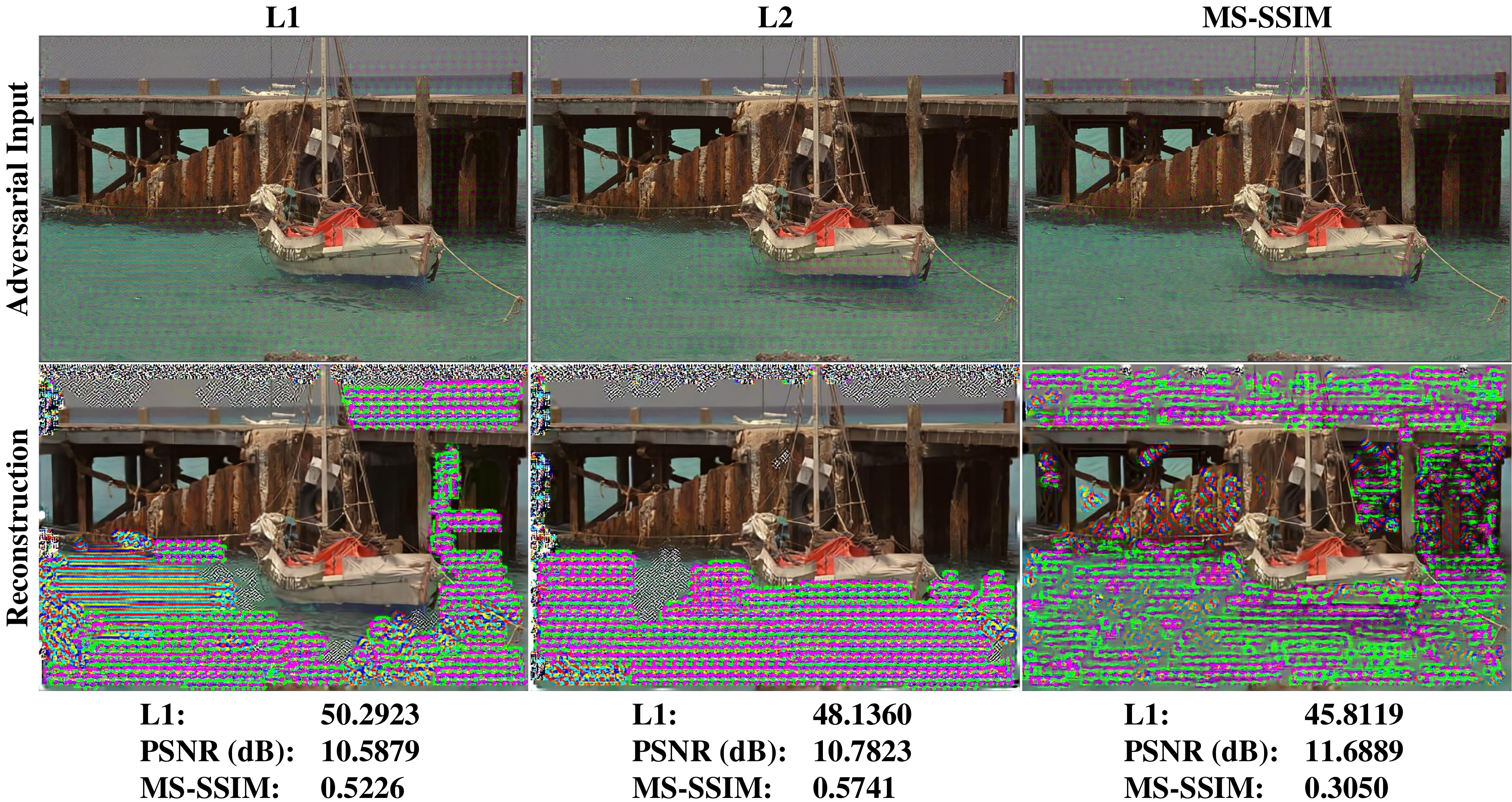}
\caption{Illustration of adversarial examples and corresponding adversarial reconstructed images using different distance measurements in~\eqref{eq:dloss}.}
\label{fig:metrics_attack}
\end{figure}

We originally apply the common $l_2$ distance in \eqref{eq:dloss} for the generation of adversarial examples. Here, we show that other distance metrics like $l_1$ distance and MS-SSIM measurement can also be used to produce adversarial examples that can also effectively attack NIC models for impaired reconstructions as shown in Fig.~\ref{fig:metrics_attack}.  

\begin{figure}[htbp]
\centering
\includegraphics[scale=0.21]{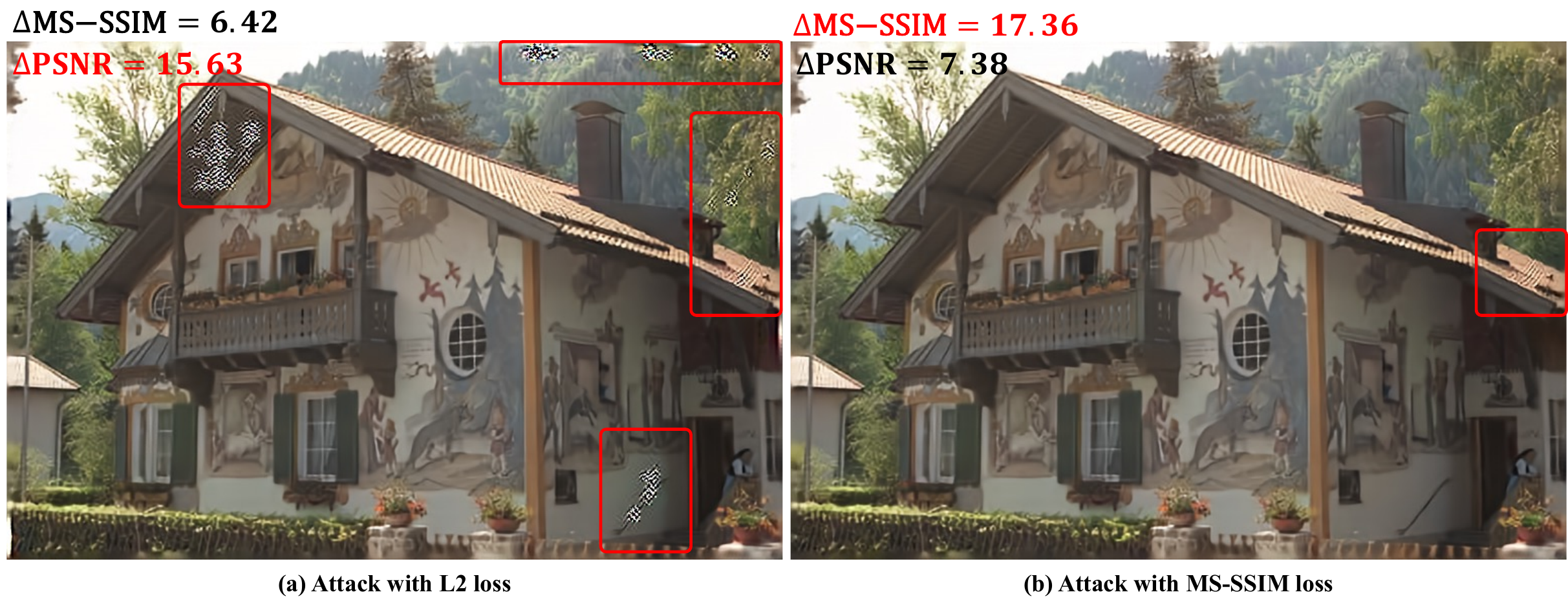}
\caption{Visual comparison of distortion using different distance measurements in~\eqref{eq:dloss}. An MS-SSIM optimized Ball\'e2018 model is tested.}
\label{fig:att_metrics}
\end{figure}

In Table~\ref{tab:att_metrics}, we provide more quantitative results. As seen, regardless of the metric used for model training, a higher $\Delta$MS-SSIM loss is observed when MS-SSIM is used as the $\mathcal{D}()$ in~\eqref{eq:dloss}; while a larger $\Delta$PSNR  is presented when $\mathcal{D}()$ uses the L2 (MSE) loss function, regardless of the loss metrics used to train the NIC models.
As illustrated in Fig.~\ref{fig:att_metrics}, although the $\Delta$MS-SSIM of Fig.~\ref{fig:att_metrics}(b) is much higher than that of Fig.~\ref{fig:att_metrics}(a), the reconstruction quality of the L2 attacked example is degraded with more visually perceivable impairments. Therefore, $\Delta$PSNR can better reflect the quality degradation incurred by the attack.

\begin{table}[htbp]
\centering
\caption{Quality evaluation comparison measured by $\Delta\textbf{PSNR}$ and $\Delta\textbf{MS-SSIM}$. Kodak datasets and Ball\'e2018 models are tested.}
\begin{tabular}{c|c|c|c}
\hline
\textbf{metrics in $\mathcal{D}$ (attack)}          & \textbf{metrics (train)} & $\Delta\textbf{PSNR}$               & $\Delta\textbf{MS-SSIM}$ \\
\hline
                                   & ms-ssim                 & 8.22                         & {\bf\color{red} 14.11}\\
\multirow{-2}{*}{\textbf{MS-SSIM}} & mse                     & 5.00                            & {\bf\color{red} 16.05}\\\hline
                                   & ms-ssim                 & {\bf\color{red} 16.45} & 5.65\\
\multirow{-2}{*}{\textbf{L2 (MSE)}}     & mse                     & {\bf\color{red} 11.38} & 9.20\\
\hline             

\end{tabular}
\label{tab:att_metrics}
\end{table}

\subsection{Performance on other types of degradations}
\begin{figure}[htbp]
    \centering
\includegraphics[width=8cm]{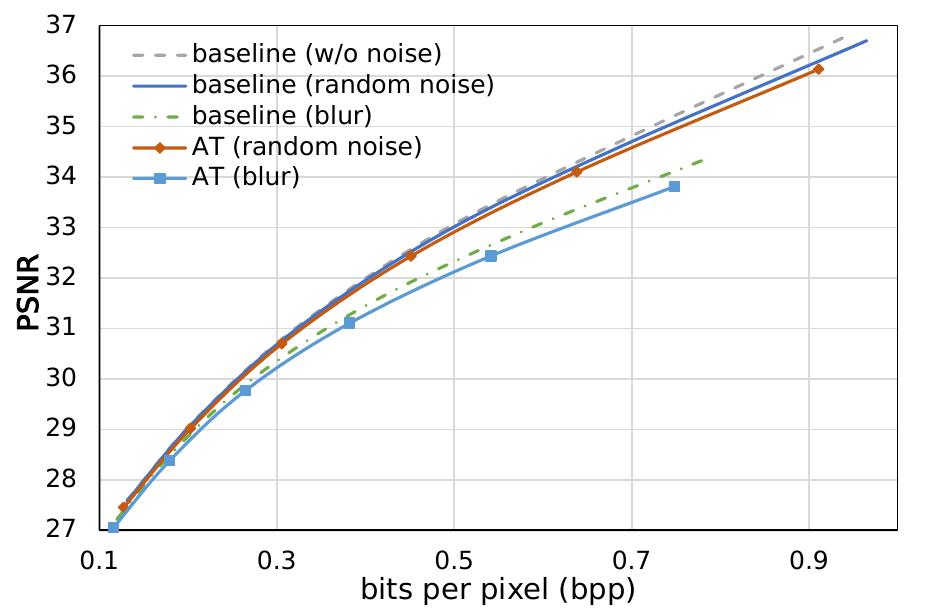}
    \caption{Rate-Distortion Comparison. Pretrained Ball\'e2018 models and corresponding AT models are tested. As shown in Fig.~\ref{fig:vi}, the random noise can be suppressed by the compression networks, making the rate-distortion performance on random noise images better than that on blurry images.}
    \label{fig:rd_degrad}
\end{figure}

We also test the rate-distortion performance of the baseline end-to-end codecs and our proposed defense method (Adversarial Training) on the noisy and blurry degradation in Fig.~\ref{fig:rd_degrad}. The results show that the AT models and baseline models show similar performance facing these scenarios. Gaussian random noise and gaussian blur are added to generate degraded input images.

\subsection{Detailed Explanations about $\Delta$PSNR}
\textbf{A Linear Example.}
In order to better clarify our reason for choosing $\Delta$PSNR, we first perform a test on a toy model, which is composed of a single-layer convolution for both the encoder $\mathsf{W_1}$ and decoder $\mathsf{W_2}$ (without non-linear activation function to ensure linearity). In this case, the network is purely linear. Therefore theoretically the amplification level for generated perturbation $\mathbf{n}_{i}$ added onto input $\mathbf{x}$ can be considered consistent with all different perturbation levels:

\begin{equation}
\begin{aligned}
\textit{with } \hat{\mathbf{x}} &= \mathsf{W}_1\mathsf{W}_2 \mathbf{x},\\
\hat{\mathbf{x}} + \mathbf{N}_{out} &= \mathsf{W}_1\mathsf{W}_2 (\mathbf{x} + \mathbf{n}_{i}) \\ &=\mathsf{W}_1\mathsf{W}_2\mathbf{x} + \mathsf{W}_1\mathsf{W}_2 \mathbf{n}_{i} \\
\textit{now we have: }\mathbf{N}_{out} &= \mathsf{W}_1\mathsf{W}_2 \mathbf{n}_{i} \\
\textit{and: } k\mathbf{N}_{out} &= \mathsf{W}_1\mathsf{W}_2 (k\mathbf{n}_{i})
\end{aligned}
\end{equation}

As shown in Fig.~\ref{subfig:linear}, the $\Delta$PSNR keeps the same for all levels over 100dB difference, while even for this linear system the VI still varies with the absolute perturbation level (the larger the absolute input perturbation, the larger the VI would be). 

\textbf{Non-linear System.}
A more complicated neural network can be very non-linear. Figure~\ref{subfig:dpsnr_nic} shows the vulnerability of a non-linear deep neural network-based image compression network. $\Delta$PSNR now also varies at different perturbation levels for this non-linear network. The results indicate that the network would greatly amplify the attack perturbation of large values. 

Note that here the VI values at large input perturbation levels ($<$ 40dB as exemplified, lower in dB means larger perturbation) become abnormal ($<$0) since the output PSNR is now negative. The output distortion is extremely large and has been out of the range of [0,1]. 

In practice, we would perform a value clipping operation to the output reconstruction to ensure it is within the range of [0-1]. With clipping operation performed, the results are shown in Fig.~\ref{subfig:dpsnr_clip}. Similar to the observations in Fig.~\ref{subfig:linear}, the VI value still gradually increases as the perturbation level increases. Besides, when the perturbation level reaches a certain threshold, both $\Delta$PSNR and VI metrics exhibit a decline. This doesn't mean that the network is less vulnerable to large perturbations. In fact, this is caused by the value clipping operation. When the input perturbation becomes significantly large, the output distortion can no longer increase due to the clipping operation.

In conclusion, our goal is to eliminate the impact of perturbation magnitude itself on the metrics. This allows us to focus on the model’s inherent vulnerability (which can be different at different perturbation levels) but would not being directly biased by the absolute perturbation level.

\subsection{Extending the Methodology to Targeted Attack}
\label{sec:targeted}
This companion section showcases that our methodology can be also applied to targeted attack. Different from the untargeted attack in previous sections that tends to just maximize the reconstruction distortion, A targeted attack is typically applied for semantic content understanding, which attempts to produce a targeted output that is designated to be different from the original output.

\subsubsection{Targeted Adversarial Examples}
Assuming that we have the source image $\bf x$ and a target image ${\bf x}^t$ that is different from $\bf x$, we hope to generate adversarial input $\bf x^*= x + \mathbf{n}$, so that the reconstructed $\hat{\bf x}^* = f_D(f_E({\bf x^*}))$ would  be close to the reconstruction of the target image $\hat{\bf x}^t = f_D(f_E({\bf x}^t))$. 
%One popular research domain of adversarial attack is targeted attack with specific target labels especially for many computer vision tasks like classification. Similarly for targeted image compression attacks, we have both the original image with embedded noise $x^*=x+\mathbf{n}$ and its corresponding reconstruction $\hat{x}^* = \mathbf{D_{\phi}}(\mathbf{E_{\theta}}(x^*))$, target image $x^t$ and its corresponding reconstruction $\hat{x}^t$. 
By fixing the encoder $f_E()$ and decoder $f_D()$ of a specific NIC method, the generation of adversarial examples for targeted attack can be formulated as minimizing the input noise as well as the output distance with the target output:
% $\lambda$ here is to control the penalty between noise and distortion.

% \begin{equation}
%     \mathop{\arg\min}_{\mathbf{n}} L_{t} =  ||\mathbf{n}||_2 + \lambda||\hat{x}^{*}, \hat{x}^{t}||_2
% \label{eq:tloss}
% \end{equation}

\begin{align}
    \mathop{\arg\min}_{\mathbf{n}}{L_{t}} =\left\{ \begin{array}{lc}
        \frac{||\mathbf{n}||_2^2}{\textbf{N}}, & \frac{||\mathbf{n}||_2^2}{\textbf{N}} \geq \epsilon \\
        \frac{||\hat{\bf x}^{*}-\hat{\bf x}^{t}||_2^2}{\textbf{N}}, & \frac{||\mathbf{n}||_2^2}{\textbf{N}} < \epsilon 
    \end{array}
    \right.
    \label{eq:tloss}
\end{align}
Similar to~\eqref{eq:dloss}, here $\epsilon$ is the threshold of the input noise to control the perturbation level.
The main difference between targeted attack in~\eqref{eq:tloss} and untargeted attack in~\eqref{eq:dloss} is that: instead of using negative $l_2$ distance that maximizes the output distortion as in~\eqref{eq:dloss}, here the objective is to minimize the distance between the reconstruction of target image $\hat{\bf x}^t$ and the reconstruction of the adversarial example $\hat{\bf x}^*$.

\begin{figure}[t]
\centering
\includegraphics[width=0.48\textwidth]{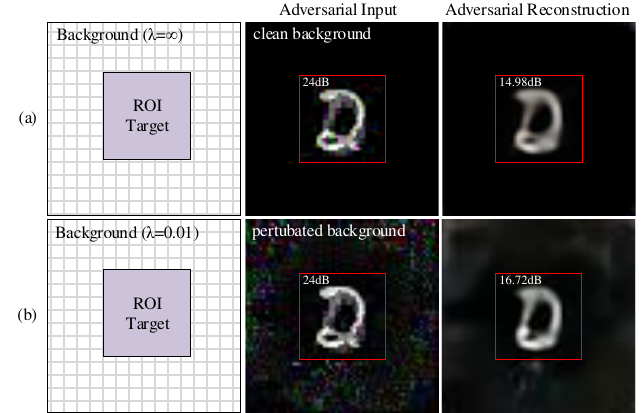}
\caption{\textbf{Attack with target area mask.} (a) example of adversarial attack with masked ROI (region of interest) target. Zero perturbation is added in the background area. (b) perturbations are added in both the target area and background area with $\lambda=0.01$ in~\eqref{eq:masked_tloss}. 
%\edit{Note that the PSNR value of adversarial input in the target area is compared with the source image while for the adversarial reconstruction, the target image is used.}
With same level of target area input perturbation, the attack efficiency is further improved by allowing some extra perturbation in the background area.}%More noise is observed on background area than on masked object shown in  adversarial example of (b).  }
\label{fig:back_roi}
\end{figure}

Note that \eqref{eq:tloss} simply hopes to minimize the distance between the source image and the target image of all pixels, which works well for low-resolution, thumbnail images, such as the handwritten digits dataset MNIST. Nowadays,  an image often exhibits a much higher spatial resolution, such as the 4K, 8K, etc, and usually presents multiple semantic cues, e.g., diverse foreground objects,  within the same scene, making it difficult to directly apply the \eqref{eq:tloss}.
%The target labels in computer vision tasks like image classification are usually binary or discrete numbers. However, for a image compression task, The target or the label becomes a full-resolution image that has the same size of the input image. It is evidently more difficult to generate adversarial examples now because the distance between the original output and target output are directly in high dimensional pixel domain. Therefore, Eq.~\ref{eq:tloss} is only suitable for images with low resolution and simple objects like MNIST. 
For most real-life images, we then suggest a more practical strategy that focuses on alerting salient object or area (e.g. text, face, etc) that is attentive to human observers. 
The target in an image usually occupies a small fraction  of  the  entire  scene, thus we can simply mask out the target area manually or by any target segmentation / saliency detection algorithms while keeping the background unchanged as in Fig.~\ref{fig:back_roi}(a).

Considering that the stacked convolution layers and resolution resampling used in most NIC solutions would enlarge the receptive field of the network, it implies that the perturbation added in the surrounding background area would also potentially affect the reconstruction of the target area. 
%\edit{Though it is possible to just perturb pixels/elements associated with the target and surrounding area,} 
To fit the general use cases. we still assume that the noise can be added to the entire image, but set different weights in target area ${\bf x}_{\mathsf{roi}}$ and background area ${\bf x}_{\mathsf{bkg}}$ in optimization, e.g.,
% \begin{equation}
%     \mathop{\arg\min}_{x^*} L = \lambda_{1} ||x_{\mathsf{roi}}, x_{\mathsf{roi}}^*||_2 + \lambda_{2} ||x_{\mathsf{bkg}}, x_{\mathsf{bkg}}^*||_2 + ||\hat x^*, \hat x^t||_2    
% \label{eq:masked_tloss}
% \end{equation}
\begin{align}
    &\mathop{\arg\min}_{\mathbf{n}}{L_{t}} \nonumber\\
    & = \left\{ \begin{array}{lc}
        \frac{||{\bf x}_{\mathsf{roi}}-{\bf x}_{\mathsf{roi}}^*||_2^2}{\textbf{N}} + \lambda\frac{||{\bf x}_{\mathsf{bkg}}-{\bf x}_{\mathsf{bkg}}^*||_2^2}{\textbf{N}}, & \frac{||\mathbf{n_\mathsf{roi}}||_2^2}{\textbf{N}} \geq \epsilon,\\
        \frac{||\hat{\bf x}_{\mathsf{roi}}^{*}-\hat{\bf x}_{\mathsf{roi}}^{t}||_2^2}{\textbf{N}} + \lambda\frac{||\hat{\bf x}_{\mathsf{bkg}}^{*}-\hat{\bf x}_{\mathsf{bkg}}^{t}||_2^2}{\textbf{N}}, & \frac{||\mathbf{n_\mathsf{roi}}||_2^2}{\textbf{N}} < \epsilon.
    \end{array}
    \right.
    \label{eq:masked_tloss}
\end{align} Here the ${\bf n}_\mathsf{roi}$ and $\epsilon$ are the noise added in the target area and its threshold. 
If $\lambda=\infty$ as in Fig.~\ref{fig:back_roi}(a), Equation~\eqref{eq:masked_tloss} is degenerated to~\eqref{eq:tloss} that no noise is allowed in the background area. Having a smaller $\lambda$ that allows some perturbation in the background area can improve the attack efficiency as shown in Fig.~\ref{fig:back_roi}(b). 

\subsubsection{Attack Evaluation}
\begin{figure}[htbp]
\centering
\subcaptionbox{MS-SSIM opt.}{\includegraphics[scale=0.9]{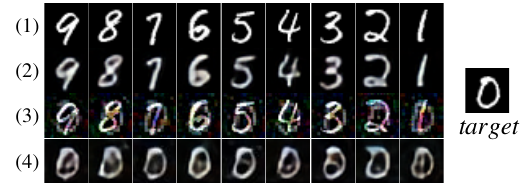}}
\par\medskip
\subcaptionbox{MSE opt.}{\includegraphics[scale=0.9]{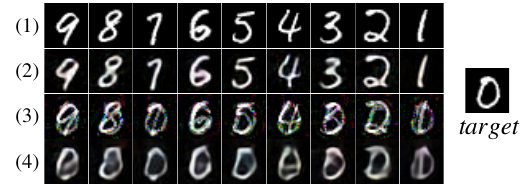}}
\caption{Targeted adversarial examples and  reconstructed output of MNIST~\protect\citeRef{lecun1998gradient} samples compressed using (a) MS-SSIM loss optimized Ball\'e2018 model, and (b) MSE loss optimized Ball\'e2018 model; (1) original source images, (2) decoded reconstructions of original images, (3) adversarial examples, (4) adversarial reconstructions. The same target image is shown on the right for both MSE and MS-SSIM optimized image compression models.}
\label{mnist_comparison}
\end{figure}

Adversarial examples generated with the optimization function described in~\eqref{eq:tloss} on MNIST dataset are shown in Fig.~\ref{mnist_comparison}. Note that original MNIST dataset are composed of single channel grayscale images. To feed them into image compression models that normally input 3-channel RGB color images, we duplicate them to 3 channels with equal values. Experiment results show that by injecting  noise perturbation to input image as in \eqref{eq:tloss}, the adversarial reconstruction significantly differs from the original input and is visually even closer to the designated target. Both MSE and MS-SSIM optimized compression models are  vulnerable to such targeted attack.

\begin{figure}[t]
\centering
\includegraphics[scale=0.45]{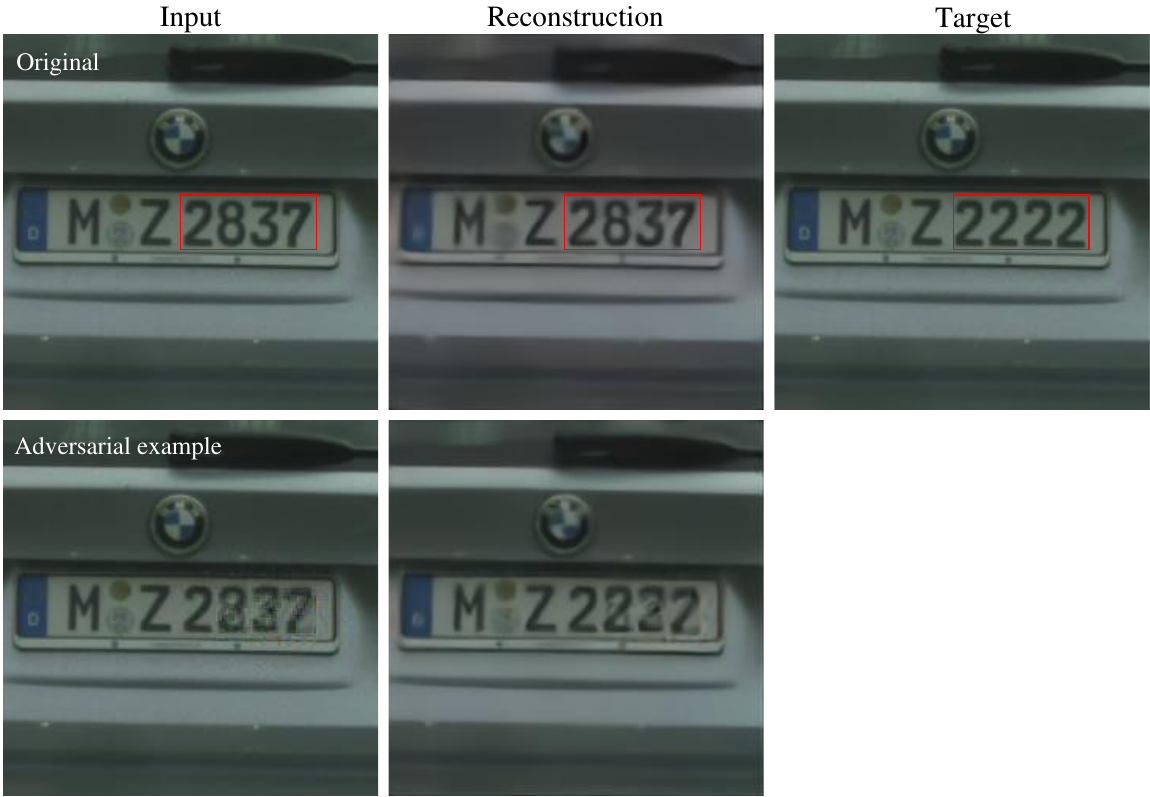}
\caption{Targeted adversarial attack on Cityscapes dataset with plate ROI mask. The license plate number area is marked in red box as the target area.}  
\label{license}
\end{figure}

A high-resolution large-scale dataset Cityscapes~\citeRef{Cordts2016Cityscapes} that contains a diverse set of stereo videos recorded in street scenes from 50 different cities, is also tested to demonstrate the effectiveness of targeted adversarial attacks with area masking on car plates.
%practical car plate detection used in autonomous driving and surveillance.  
We apply \eqref{eq:masked_tloss} to generate targeted adversarial examples where we set $\lambda = 0.1$. %\edit{and run 10000 steps with learning rate at 1e-3 and the learning rate is decayed by half for each 3333 steps.} 
%according to~\eqref{eq:masked_tloss} on more practical scenarios like autonomous self-driving and surveillance. $\lambda = 0.25$ with 10000 steps iteration for targeted attack with learning rate equals 1e-3 and the learning rate is decayed by half each 3333 steps. 
%With approximately the same level of input noise added onto the input image, the reconstruction distortion in the target area compared with the target image during adversarial example generation shown in Fig.~\ref{fig:rec_loss} can be further reduced with the target area masking, proving the effectiveness of this strategy. 
As shown in Fig.~\ref{license}, the numbers in the car plate can be successfully modified to have the adversarial reconstruction closer to the designated target, which may consequently affect the inference engine for decision-making in autonomous driving and surveillance.

\subsubsection{Application Limitations}
First, similar to the defense strategy in Sec.~\ref{sec:attack_defense}, we can also include targeted adversarial examples into the training stage to finetune the model for improved robustness. However, in real-life applications, the ``target'' of adversarial attack in pixel domain is abundant, making it impossible to include all potential ``targets''. A more practical solution for the defense of targeted attack is to limit the methodology developed in this work to some specific task.

Besides, we mainly focus on pixel-wise attack that affects the reconstruction visually in this paper. Another important domain of adversarial attack is to fool the artificial system (e.g. the accuracy of face recognition system) instead of human eyes. We leave the problem of the defense of targeted attack and the attack of NIC solutions in computer vision tasks for future study.

\bibliographystyleRef{IEEETran}
\bibliographyRef{main}

\end{document}